\documentclass{article}

\usepackage{PRIMEarxiv}

\usepackage[utf8]{inputenc} 
\usepackage[T1]{fontenc}    
\usepackage{hyperref}       
\usepackage{url}            
\usepackage{booktabs}       
\usepackage{amsfonts}       
\usepackage{nicefrac}       
\usepackage{microtype}      
\usepackage{lipsum}
\usepackage{fancyhdr}       
\usepackage{graphicx}       
\usepackage{amsmath}
\usepackage{siunitx}
\usepackage{authblk}        
\graphicspath{{media/}}     

\title{Magnetic Tactile-Driven Soft Actuator for Intelligent Grasping and Firmness Evaluation
\thanks{This work was supported by the RAISE Project “Robotics and AI for Socio-economic Empowerment” (ECS\_00000035 J33C22001220001).}
}

\author[ 1,2,3,4]{Chengjin Du}
\author[ 2,4]{Federico Bernabei}
\author[ 2,5]{Zhengyin Du}
\author[ 2]{Sergio Decherchi}
\author[ 6]{Matteo Lo Preti}
\author[ 2*]{Lucia Beccai}

\affil[1]{Shien-Ming Wu School of Intelligent Engineering, South China University of Technology, Guangzhou, China}
\affil[2]{Soft BioRobotics Perception Lab, Istituto Italiano di Tecnologia (IIT), Genova, Italy}
\affil[3]{The BioRobotics Institute, Scuola Superiore Sant’Anna (SSSA), Pisa, Italy}
\affil[4]{Department of Haptic Intelligence, CVTE Guangzhou Shiyuan Electronics Co., Ltd, Guangzhou, China}
\affil[5]{Università degli Studi di Genova (UniGe), Genova, Italy}
\affil[6]{Department of Mechanical Engineering, National University of Singapore (NUS), Singapore}

\begin{document}
\maketitle

\begingroup
  \renewcommand\thefootnote{*}
  \footnotetext{Corresponding author:\texttt{lucia.beccai@iit.it}}
\endgroup

\vspace{-1em}
\begin{abstract}
\noindent\hspace*{1em}Soft robots are emerging as powerful tools for manipulating delicate objects and interacting with complex environments, but their adoption is hindered by two critical gaps: the lack of fully integrated tactile sensing and the distortion of sensor signals caused by actuator deformations. This paper addresses both challenges by introducing the SoftMag actuator: a magnetic tactile-sensorized soft actuator. Unlike previous systems that rely on attached sensors or treat sensing and actuation separately, SoftMag unifies them through a shared design architecture while explicitly confronting a long-overlooked issue: the mechanical parasitic effect, where actuator-induced deformations corrupt tactile signals. A multiphysics simulation framework is developed to model this coupling, and a neural-network-based decoupling strategy is proposed to isolate and remove the parasitic component, restoring sensing fidelity. Comprehensive experiments including indentation, quasi-static and step actuation, and fatigue tests validate the actuator’s performance, durability, and decoupling effectiveness. Building upon this actuator-level foundation, the system is extended into a two-finger SoftMag gripper, where a multi-task neural network enables real-time prediction of tri-axial contact forces and contact position. Furthermore, a probing-based strategy is introduced to estimate object firmness during grasping. Validation on apricots demonstrates a strong correlation (Pearson $r>0.8$) between gripper-estimated firmness and reference indentation measurements, confirming the system’s capability for non-destructive, in-process quality assessment. The results demonstrate how combining integrated magnetic tactile sensing, learning-based signal correction, and real-time inference enables a soft robotic platform that not only adapts its grasp but also quantifies material properties during grasping. While fruit handling serves as an application case, the presented framework offers an approach for advancing sensorized soft actuators toward intelligent, material-aware robotics.
\end{abstract}


\section{Introduction}
\noindent\hspace*{1em}In recent years, soft robotics has witnessed remarkable advances in developing sensorized actuators that blend the inherent compliance of soft materials with high-fidelity tactile sensing \cite{kar2022review,qu2023recent}. By integrating tactile feedback into soft actuators, these systems are able to mimic some of the subtle sensing capabilities found in biological systems \cite{li2019bio,kang2024biomimic}, offering the potential of performing delicate manipulation tasks in non-visual dynamic environments \cite{xu2024soft,costi2022environment}. Current approaches often focus on simply “attaching” existing (or commercial) sensors to soft actuators rather than developing a genuinely integrated solution. The former strategy lacks a systematic framework for the design, characterization, and evaluation of sensorized soft actuators, leading to inconsistencies in behavior and difficulties in assessing performance \cite{subad2021soft,gunderman2022tendon}. Meanwhile, the integration of magnetic-based tactile sensors into soft grippers remains unexplored, despite showing already great promise for tactile sensing in highly soft agents \cite{wang2016design,yan2021soft}. Moreover, integrating tactile sensors into soft actuators presents a unique and yet unavoidable challenge, namely dealing with the mechanical parasitic effects arising from sensor-actuator coupling. This effect, where sensor readings are distorted by the actuator's inherent deformations and motions, can significantly compromise the accuracy of tactile feedback during rapid or complex maneuvers, thereby limiting the overall system performance \cite{hegde2023sensing, faris2023proprioception}\\
\noindent\hspace*{1em}Beyond these general challenges, the integration of tactile sensors directly into soft gripper opens up transformative possibilities in application domains such as fruit handling \cite{bac2014harvesting,zujevs2015trends}. In agriculture, tasks such as sorting, packaging and quality monitoring are often handled separately from ripeness or firmness evaluation, leading to hardware redundancy, increased complexity, and inefficiencies in post-harvest operations \cite{bhargava2021fruits,gao2009automatic,shipman2021can}. Among various quality indicators such as size, color, and surface defects, firmness stands out as a critical parameter that reflects internal tissue structure and ripeness stage. This is especially true for climacteric fruits like avocados, kiwis, and peaches, which soften progressively due to enzymatic degradation and moisture redistribution \cite{StableMicroSystems2025,FreshKnowledge2025,kader2002postharvest,landahl2020non}. Traditional firmness assessment relies on destructive methods such as the Magness-Taylor (MT) puncture test, which, while accurate, damages the fruit and is unsuitable for large-scale or in-line deployment \cite{ding2024non,abbott1976effe}. Non-destructive alternatives, including acoustic response \cite{Vanoli2012}, low-force indentation \cite{fathizadeh2021nondestructive}, and near-infrared (NIR) spectroscopy \cite{lupu2022techniques}, have been proposed, but these often suffer from environmental sensitivity and require separate inspection stages that hinder full system integration \cite{zhou2022rapid,lakshmi2017non}. Soft robotic grippers offer a compelling solution by enabling gentle, damage-free grasping and providing a platform for embedded sensing. However, existing systems typically treat manipulation and sensing as decoupled processes, or rely on offline, categorical classification rather than real-time quantitative assessment \cite{rizzo2023fruit,moggia2022comparison}. These limitations point to a critical gap: the absence of a unified, non-destructive system capable of both grasping and evaluating fruit firmness in a single operation.\\
\noindent\hspace*{1em}Motivated by these gaps, this work attempts to address not only the fundamental requirements of soft tactile-driven grasping but also extends these solutions to the specialized demands of fruit handling. By establishing a unified framework for developing sensorized soft grippers and introducing a compact data-driven approach to mitigate mechanical parasitic effects, the proposed system offers reliable tactile-driven grasping for general tasks as well as real-time, quantitative firmness evaluation specifically tailored to fruit handling needs. In support of this data-driven methodology, a multi-task neural network is employed for tactile inference. This combination improves the gripper’s effectiveness in manipulating delicate objects, such as fruits, while simultaneously enabling firmness assessment within a single operation. \\
\noindent\hspace{1em}The key contribution of this work consists of introducing a novel magnetic tactile-sensorized soft actuator (SoftMag actuator), and a tactile-driven soft gripper (SoftMag gripper) capable of performing adaptive grasping. In this framework, the mechanical parasitic effect that prevails (yet often overlooked) for tactile-driven sensorized actuators and grippers is addressed. Moreover, by exploiting the SoftMag gripper, a real-time quantitative firmness estimation by soft probing is provided as part of the fruit handling process.\\
\noindent\hspace*{1em}The remainder of the manuscript is organized as follows. Section 2 provides a review of the state of the art in soft grippers, tactile sensing, and sensorized soft gripper systems, with particular focus on applications in fruit handling. Section 3 introduces the development of the SoftMag actuator, detailing its design, fabrication, and performance. Section 4 presents the integration of the actuator into the two-finger SoftMag gripper, together with its grasping capabilities and robustness to magnetic interference. Section 4.4 describes a multi-task learning framework for predicting force and contact information from magnetic signals. Section 5 addresses the issue of mechanical parasitic effects through a neural-network-based decoupling method. Section 6 introduces the probing-based strategy for real-time firmness evaluation during grasping, with experimental validation on diverse objects and fruits under both progressive and individual testing protocols. Section 7 discusses the broader implications of this work, outlines current limitations, and suggests future research directions. Finally, Section 8 concludes the manuscript with a summary of contributions and reflections on potential industrial deployment.

\section{Related Works}
\noindent\hspace*{1em}Soft robotic grippers have gained substantial attention for their ability to handle diverse objects through the inherent compliance of soft materials, which enables passive adaptation to various shapes and fragilities. This adaptability proves especially valuable in scenes where object safety and mechanical versatility are paramount, distinguishing soft grippers from traditional rigid systems \cite{shintake2018soft,abozaid2024soft,qu2024advanced}. At the same time, tactile sensing is integral to optimizing grasping performance, as it provides critical data on contact forces, slip events, and object textures \cite{kim2022soft,mandil2023tactile}. Recent advances in sensor technologies relevant to soft robotics have led to a broad array of sensing modalities including resistive, capacitive, optical, visual-based, and magnetic sensors \cite{meribout2024tactile,chi2018recent,wang2023tactile}, offering high-resolution data acquisition for force estimation and object classification, allowing soft grippers to adjust their grasp in real time \cite{roberts2021soft, alshawabkeh2023highly}. Nevertheless, integrating tactile sensors within soft actuators presents significant challenges. The prevailing approach in current research primarily involves retrofitting commercial sensors into soft actuators rather than developing fully integrated sensing solutions. This reactive strategy often leads to mechanical interference and constrained design flexibility, ultimately limiting the effectiveness of tactile feedback \cite{russo2017design,kultongkham2021design}. Meanwhile, although magnetic-based tactile sensors have demonstrated considerable potential, their application in soft grippers remains unexplored \cite{yan2024soft,hu2024large}.
\subsection{Tactile-Driven Soft Grippers}
\noindent\hspace*{1em}Recent efforts have various strategies for integrating sensors into soft grippers to enable adaptive grasping in real-world scenarios, improving object stability and enabling classification based on mechanical properties. In many of these works, commercial force sensors are retrofitted into soft actuators to expedite development. For example, Liu\cite{liu2024soft} introduced a soft gripper with direct attachment of commercialized force and bend sensors for slip detection and shape adaptation. Hedge\cite{hegde20243d} developed a 3D-printed mechano-optic force sensor with high sensitivity, while failing to address actuation-induced distortions. Low\cite{low2021sensorized} likewise proposed a sensorized reconfigurable soft gripper, but, as with other examples, the direct presence of external or rigidly mounted sensors may compromise overall compliance and adaptability. Some studies have attempted customized or fully integrated sensing solutions. For instance, Zhang\cite{Zhang2024} developed a customized integrated sensing solution by separating bending and contact detection, but it remains to be proven that the two sensing modalities are fully independent. Zhao\cite{zhao2016optoelectronically} developed a prosthetic hand integrated with stretchable multimode optical waveguides for proprioception and exteroception, demonstrating a basic firmness classification function. Nonetheless, the design and fabrication of a single finger are rather complicated, and the experiment is preliminary, limited by the small volume of the tested samples. Moreover, these approaches continue to grapple with mechanical parasitic effects and lack an integrated design and characterization framework, highlighting the need for simpler yet robust and systematic solutions.
\noindent\hspace*{1em}To interpret the often non-linear or high-dimensional tactile data generated by soft actuators, artificial intelligence (AI) methods have been adopted by many studies. For example, Zhang\cite{zhang2024ai} integrated piezoresistive sensors into a soft robotic gripper and employed a Swin-Transformer-network-co-pilot algorithm to achieve high-accuracy object recognition. Deng (2022) introduced a liquid metal (EGaIn) sensor for force, stiffness, and deformation detection, combining it with a convolutional neural network for object classification. Zimmer\cite{Zimmer2019} studied the performance of different AI models for improving grasp stability. While these methods have merit in reducing the modeling workload, their deployment and performance in real-time soft gripper systems remain rarely discussed. On the other hand, active grasping control has become an essential strategy for managing dynamic interactions. Wang\cite{Wang2024} employed a commercialized force sensor matrix with a Kalman-filtered PID controller to realize precise contact force control. Low\cite{low2021sensorized} utilized a force-feedback loop to optimize gripping configurations. Beyond these mainstream solutions, researchers are exploring novel sensing solutions for soft grippers. Both Li\cite{Li2023} and Zhang\cite{Zhang2022a} proposed triboelectric sensor-integrated grippers for object recognition; Chen\cite{Chen2020} introduced a self-powered triboelectric nanogenerator-based sensor and integrated it on a soft gripper for preliminary grasping; Han\cite{Han2022} developed a wireless soft gripper with a graphene oxide/polyimide composite-based tactile layer; Zuo\cite{Zuo2021} utilized anti-freezing ionic hydrogel for capacitive tactile sensing in low-temperature environments. While novel, these solutions are mostly in preliminary phases or are developed for special application scenarios. Finally, as a popular solution, visual-based tactile sensors are increasingly adopted: Zhang\cite{Zhang2024} introduced TacPalm, a tactile gripper integrated with a visual-based sensor for grip adjustment; James\cite{James2020} proposed a slip-detection robotic hand using TacTip sensors and a JeVois Machine Vision System. While offering unprecedented spatial resolution, this technique lacks direct force sensing modalities and will introduce structural stiffness that conflicts with the soft actuator’s compliance. \\
\noindent\hspace*{1em}Overall, this analysis underlines persistent challenges in accuracy, response time, and robustness. In addition to the integration hurdles, the absence of an integrated characterization framework results in inconsistent evaluations of sensing performance, complicating comparisons between sensorized soft grippers\cite{Dou2021,Butt2018}. Also, as earlier mentioned, the compliant nature of soft materials and actuation-induced deformations interfere with sensor readings, leading to distorted tactile signals. This effect becomes especially problematic in dynamic or high-speed tasks, making it difficult to distinguish between true contact forces and actuation-induced artifacts\cite{Wang2024,Shu2023}. Collectively, these issues highlight the need for improved sensing integration strategies that enhance reliability without sacrificing the core advantages of soft robotic systems\cite{Yang2020}.\\
\subsection{Soft Grippers for Fruit Handling and Evaluation}
\noindent\hspace*{1em}Among specialized applications, the agricultural and food industries present especially demanding requirements\cite{Lehnert2017}. Fruit handling poses significant challenges due to wide variations in size, shape, and mechanical properties across different fruits, necessitating a gentle yet accurate approach\cite{Zhou2022}. Generally, the existing grippers are aided by vision-based sensing techniques and contact-based techniques to aid damage-free grasping. Moreover, most of the gripping solutions target the harvesting phase as cited hereafter. Wang\cite{Wang2023} developed a LiDAR-RGB-based gripper for apple harvesting, achieving a $98.2\%$ fruit identification rate. Cheng\cite{Cheng2022} introduced a vision-aided pneumatic gripper able to adapt to various fruit shapes and sizes. Filho\cite{ValeFilho2024} investigated computer vision to assist a fin ray gripper, reporting an average of $98.31\%$ recognition accuracy on orange. Visentin\cite{Visentin2023} incorporated a vision-based tactile sensor into a strawberry-harvesting gripper for enhancing grip control and reducing fruit damage. Within contact-based solutions, Cook \cite{Cook2020} integrated capacitive tactile sensors into a soft gripper but only conducted preliminary testing without further development. Gunderman\cite{gunderman2022tendon} developed a tendon-driven gripper with force feedback for blackberry harvesting, achieving a reduced damage rate. Russo\cite{russo2017design} force-sensitive resistors (FSRs) on a three-finger gripper to achieve gentler handling. Finally, the two approaches can merge, like in Navas\cite{Navas2024} where a vision-assisted soft gripper was equipped with air pressure and infrared sensors for real-time contact estimation in small fruit harvesting.\\ 
\noindent\hspace*{1em}Among the applications of existing sensing solutions, fruit quality evaluation—especially firmness—remains a key challenge. For items such as apricots, avocados, kiwis, plums, and peaches, firmness is strongly correlated with ripeness and must be assessed quickly and accurately to prevent spoilage or suboptimal packaging\cite{marc2023overview}. Traditionally, destructive approaches like the Magness-Taylor puncture test are employed, while damaging the fruit are unsuitable for in-line or high-throughput operations. To overcome this, non-destructive methods such as acoustic response, low-force indentation, and near-infrared spectroscopy have been explored. However, these alternatives are often sensitive to environmental and surface conditions and typically require additional inspection stages, preventing seamless integration into automated sorting workflows. On the other hand, most systems continue to separate grasping from fruit evaluation, often relying on offline categorical classification, external sensing, or dedicated inspection hardware. As a result, existing pipelines lack real-time, quantitative assessment capabilities during the grasping process itself. By contrast, a sensorized soft gripper capable of probing the fruit and informing about its firmness offers an opportunity to unify grasping and quality inspection into a single compact operation. This approach aligns with the increasing demand for automated, real-time, and scalable post-harvest solutions that are both efficient and gentle. To this end, Almanzor\cite{Almanzor2022} attached an electrical impedance sensor to a soft gripper, employing machine learning for banana ripeness classification. Qiu\cite{Qiu2023} integrated a near-infrared sensor into the palm of a tendon-driven gripper to detect blackberry ripeness. Xia\cite{Xia2022} demonstrated a flexible dual-mechanism pressure sensor on a rigid gripper for avocado firmness detection, while Zhang\cite{Zhang2022} and Qin\cite{Qin2024} both employed nano-piezoresistive sensors on the same gripper for evaluating the firmness of avocado and kiwi. Although these systems incorporate basic force/tactile sensing capabilities to classify fruits by ripeness, most of them provide only offline, categorical gradings, rather than real-time quantifiable measurements\cite{Pourdarbani2020,Tian2022,Zhu2021}. Moreover, the grasping process is often separated from firmness evaluation, requiring additional stations or manual intervention for quality inspection\cite{Wang2023a}. Such a fragmented workflow can lead to redundant hardware, greater operational complexity, and avoidable delays in post-harvest procedures. By contrast, a sensorized soft gripper capable of direct, quantifiable firmness evaluation would integrate the handling and ripeness assessment steps into a single, unified process, streamlining operations and potentially improving overall efficiency.

\section{Development of the SoftMag Actuator}
\label{sec:headings}

\subsection{Concept of the SoftMag Actuator}
\begin{figure}[h!]
    \centering
    \includegraphics[width=0.7\linewidth]{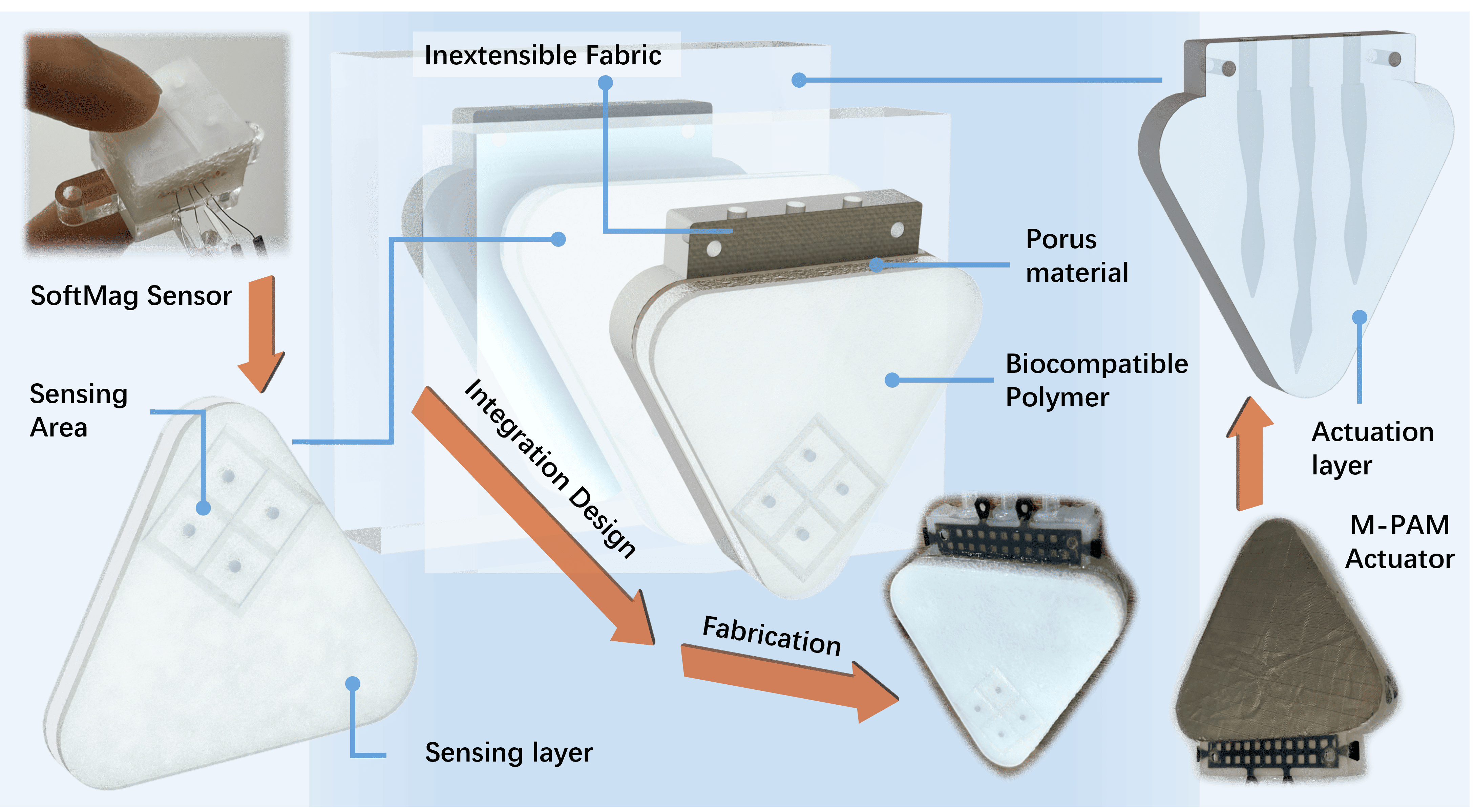}
    \caption{Concept of the SoftMag actuator: Sensorizing the M-PAM actuator with the SoftMag sensor through a soft, integrated design enabled by a shared porous material.}
  \label{fig:1.SoftMagConcept}
\end{figure}

\noindent\hspace*{1em}As illustrated in Figure~\ref{fig:1.SoftMagConcept}, the proposed sensorized SoftMag actuator is developed by integrating the previously introduced SoftMag sensor\cite{du2023design} with a customized soft pneumatic actuator M-PAM\cite{bernabei2023development} to form a unified, sensorized structure. The novelty lies in two aspects: first is the use of a magnetic-based tactile sensor in a soft actuator; second is the smooth integration of the sensing and actuation layers through shared materials and fabrication techniques. This co-design approach enhances mechanical coherence, ensures consistent actuation and sensing behavior, and simplifies assembly.\\
\noindent\hspace*{1em}The SoftMag actuator comprises two layers: a sensing layer and an actuation layer. At the core of the integration is the shared use of Soma Foama™ 15 (Smooth-On, Inc., U.S.A.), a two-part platinum-catalyzed silicone foam characterized by low density and high compressibility. This material not only provides a deformable body for tactile sensing but also enables tight coupling between the sensing and actuation layers. The sensing layer is constructed by first encapsulating four small permanent magnets within Ecoflex™ 00-30 (Smooth-On Inc., USA), which are then embedded into the porous Soma Foama™ body. This configuration yields a compliant magnetic sensing module with strong deformation sensitivity. The actuation layer consists of a trident-shaped, multi-channel pneumatic chamber also cast by Ecoflex™ 00-30, chosen for its good stretchability and softness. A laser-cut inextensible fabric (Holland Shielding Systems B.V., Netherlands) is bonded to the back of the structure to constrain unwanted expansion and guide the bending motion during pressurization. The sensing and actuation layers are then bonded using Sil-Poxy™ adhesive (Smooth-On Inc., USA), resulting in a compact, soft actuator with embedded tactile sensing capabilities at its tip. This integrated unit forms the basis of the SoftMag gripper system.\\

\subsection{Simulation Validation}
\begin{figure}[h!]
        \centering
        \includegraphics[width=0.7\linewidth]{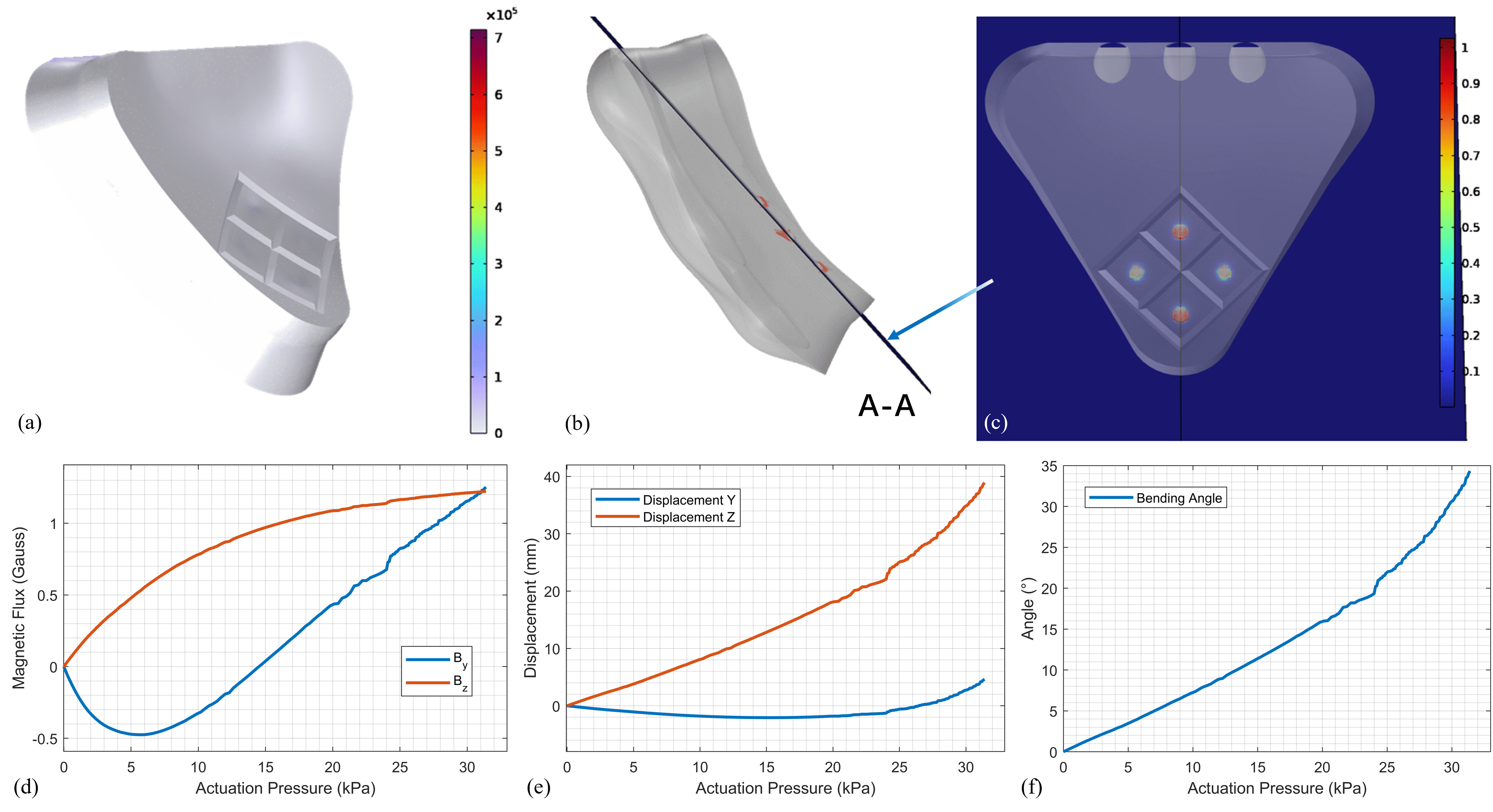}
        \caption{Results of the multiphysics simulation validation by applying a $0-31.4$ kPa pressure to the SoftMag actuator: (a) Stress and deformation under pressure load; (b) Side view and the A-A cross-section plane; (c) Magnetic flux density map on the A-A plane; (d) $\Delta B_y$ and $\Delta B_z$ at the sensing locus; (e) Tip displacement in the Y and Z directions; (f) Bending angle of the actuator.}
        \label{fig:2.Sim}
\end{figure}
\noindent\hspace*{1em}To evaluate the SoftMag actuator, a multiphysics simulation was conducted in COMSOL Multiphysics by coupling Solid Mechanics and Electromagnetics modules. The CAD geometry was imported from SolidWorks via LiveLink, and material properties were assigned according to (Marc, 2023). Structural symmetry and an overset mesh strategy were employed to enhance computational efficiency and numerical stability. The actuator was fixed at its base, and internal pressure was applied to the pneumatic chambers to simulate actuation. To reduce computation cost and improve numerical stability and accuracy, an overset mesh strategy was adopted to perform the simulation. Details about the adopted simulation framework and method can be found in Appendix A. Figure~\ref{fig:2.Sim} summarizes the simulation results over a pressure range from 0 to $~30kPa$. Particularly, magnetic field variations along the Y and Z axes shown in Figure~\ref{fig:2.Sim} (d) revealed that internal deformation influences sensor output even in the absence of external contact. This mechanical parasitic effect arises from internal actuator deformation (such as bending or compression) that induces relative displacement between the embedded magnets and the Hall-effect sensor. These internal shifts produce signal changes unrelated to contact, thereby distorting the output and compromising sensing accuracy. This phenomenon was experimentally validated (Section 3.5.1) and later addressed through a decoupling strategy. The simulation’s bending behavior was further quantified using a bending angle calculation method detailed in Appendices A, and a simulation animation is provided in Video 1.

\subsection{Fabrication Process and Materials}
\begin{figure}[h!]
    \centering
    \includegraphics[width=0.75\linewidth]{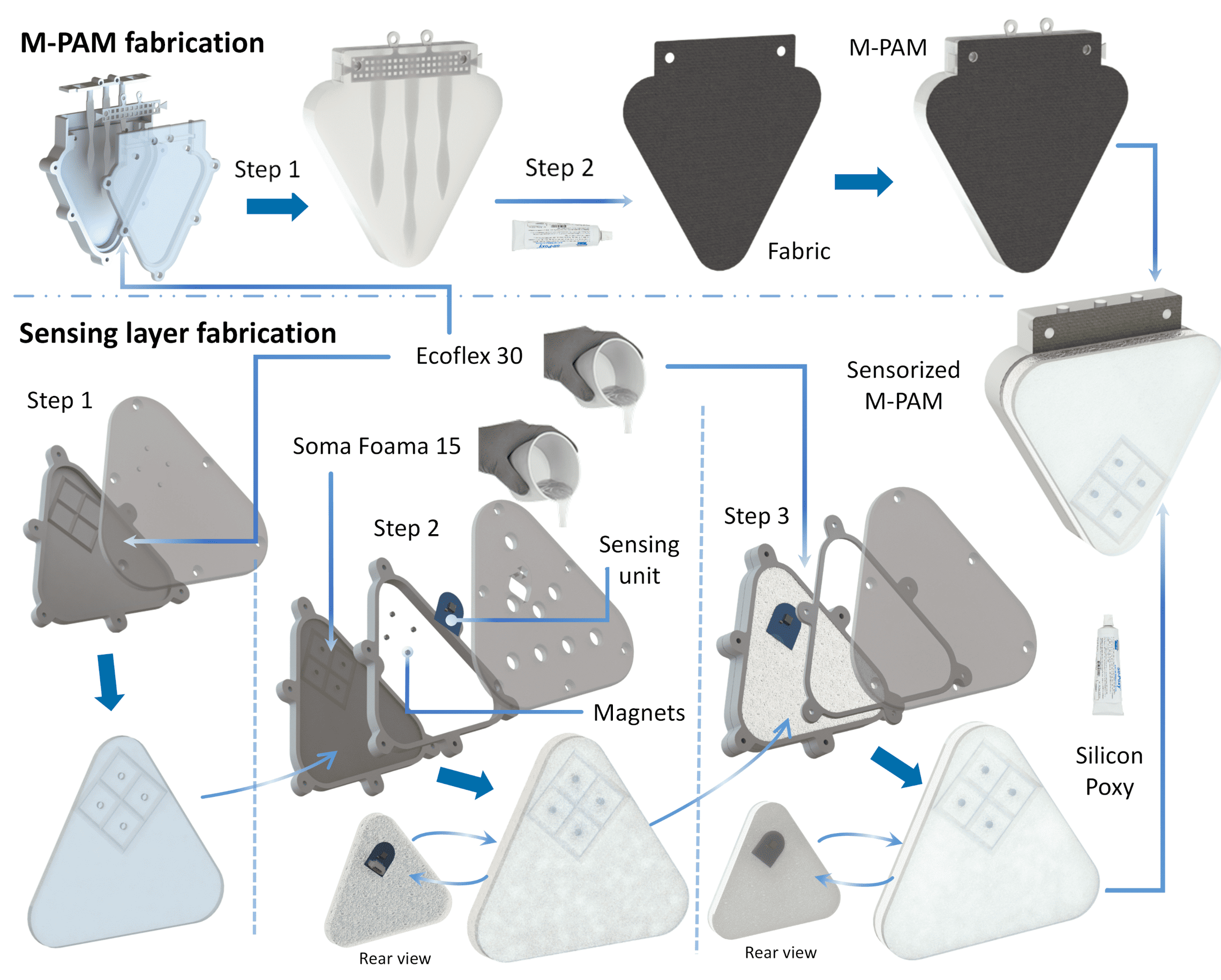}
    \caption{Fabrication process of the SoftMag actuator.}
    \label{fig:3.Fabreication}
\end{figure}
\noindent\hspace*{1em}The fabrication process of the SoftMag actuator involves two primary components: the actuation layer and the sensing layer. The actuation layer was fabricated through a two-step procedure depicted in the upper section of Figure~\ref{fig:3.Fabreication} (a): The actuation layer was fabricated using a three-part mold to form the multi-channel pneumatic chamber. After casting silicone rubber into the assembled mold, an inextensible layer was applied to the back surface to constrain expansion and guide bending. The two layers were then bonded using silicone adhesive to complete the actuation module. The sensing layer was fabricated through a two-step foam casting process. First, a base layer was cast to embed four permanent magnets, with size and orientation identical to the configuration reported in \cite{du2023design}. A second foam layer was then added to encapsulate the magnets. A Hall-effect sensor was subsequently embedded in a soft silicone layer and bonded to the foam structure, forming the sensing module. Finally, the actuation and sensing layers were adhered together to create the complete sensorized SoftMag actuator. Detailed information on mold design, 3D printing parameters, surface treatments, degassing procedures, and material specifications is provided in Appendix B.

\subsection{Sensing Performance}
\subsubsection{Indentation Test Setup and Data Processing}
\begin{figure}[h!]
    \centering
    \includegraphics[width=0.75\linewidth]{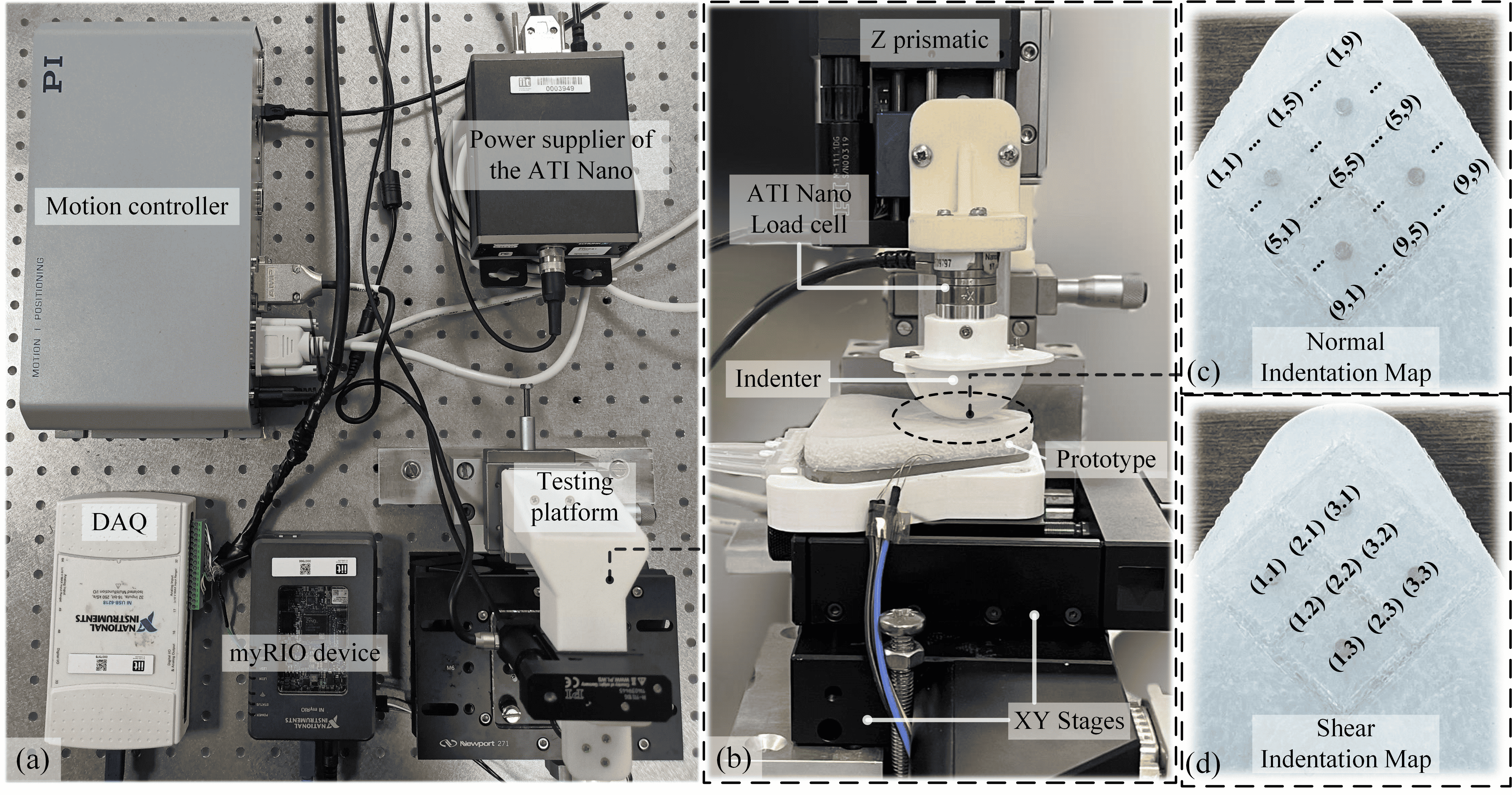}
    \caption{Indentation testing setup and protocol: (a) Overall testing setup; (b) Close-up of the indentation platform; (c) Grid map of $9×9$ indentation points used for normal indentation tests; (d) Grid map of $9×9$ indentation points used for shear indentation tests.}
    \label{fig:4.Setup}
\end{figure}

\noindent\hspace*{1em}To evaluate the SoftMag actuator’s sensing behavior, two sensorized prototypes were tested using the experimental setup shown in Figure~\ref{fig:4.Setup} (a–b). A 3D-printed acrylonitrile butadiene styrene (ABS) half-sphere probe (15 mm radius) was used for both normal and shear indentation tests. For normal tests, a 9 × 9 grid of indentation points (2 mm spacing) was tested, with 16 cyclic indentations per point, as illustrated in Figure~\ref{fig:4.Setup} (c). Shear testing was conducted at 3 × 3 positions (Figure ~\ref{fig:4.Setup}d), using cyclic ray-like translations under fixed vertical compression to simulate multi-directional shear interactions. All motion stages were manually or programmatically controlled to ensure repeatability and spatial accuracy. All normal and shear indentation datasets were processed using a standardized pipeline involving low-pass filtering (Chebyshev Type I), peak-trough segmentation, and offset correction. Detailed procedures, including the IIR filter parameters, cyclic segmentation, and noise rejection criteria, are described in Appendix B.

\subsubsection{Sensor Response and Analysis}
\begin{figure}[h!]
    \centering
    \includegraphics[width=0.75\linewidth]{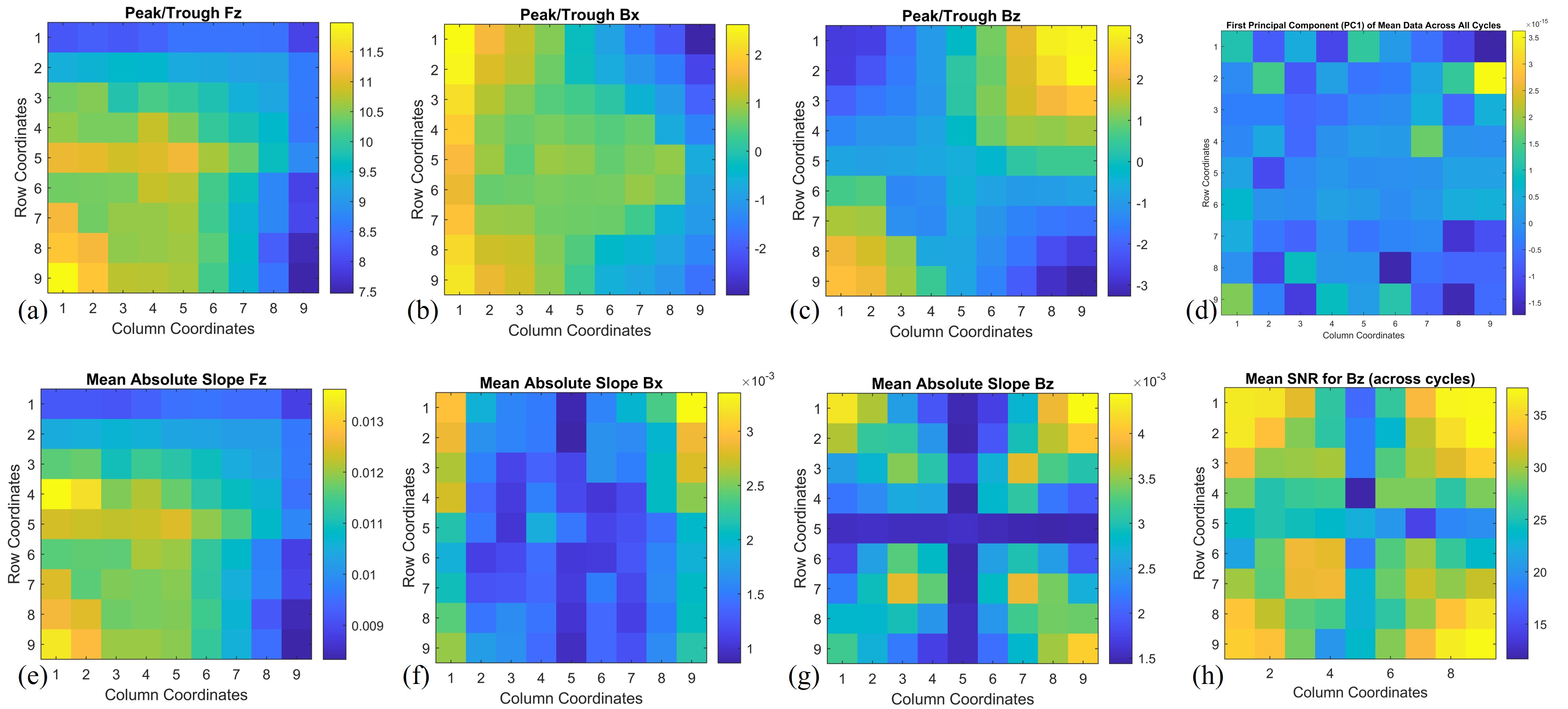}
    \caption{Spatial analysis of sensing performance during the normal indentation test: (a–c) Global distributions of peaks and troughs in normal force, $B_x$, and $B_z$ magnetic flux responses; (d) Global distribution of PC1 scores from principal component analysis of magnetic flux responses; (e–g) Global slope distributions (first derivatives) of normal force, $B_x$, and $B_z$; (h) Global mean signal-to-noise ratio (SNR) map for $B_z$ magnetic flux under unloaded conditions.}
    \label{fig:5.SpaticalAna}
\end{figure}
\noindent\hspace*{1em}To analyze the sensing behavior of the SoftMag actuator, average responses from cyclic normal indentation tests were analyzed using peak/trough extraction, slope analysis, principal component analysis (PCA), and signal-to-noise ratio (SNR) mapping. As shown in Figure~\ref{fig:5.SpaticalAna} (a–c), the global distributions of peak and trough values in normal force ($F_z$) and magnetic flux ($B_x, B_z$) reveal generally symmetric stiffness and response patterns across the sensor surface, with local deviations likely due to fabrication asymmetries. The $B_x$, and $B_z$ responses exhibit spatial distributions consistent with the sensor’s magnet configuration and mechanical design. Figure~\ref{fig:5.SpaticalAna} (e–g) shows the corresponding slope maps, further highlighting peripheral sensitivity increases due to magnet displacement and tilt under edge loading. PCA results in Figure~\ref{fig:5.SpaticalAna} (d) indicate dominant variance along the sensor’s outer regions, suggesting effective position discriminability where magnetic interference is minimal. Finally, Figure~\ref{fig:5.SpaticalAna} (h) presents the SNR map for $B_z$, revealing reduced signal fidelity near the central junction—attributed to flux coupling—while outer regions maintain higher SNR. Additional spatial maps of magnetic flux components, including peak-trough, slope, and SNR distributions, are provided in Appendix C. These results collectively confirm that the SoftMag actuator exhibits spatially consistent, mechanically coherent magnetic sensing suitable for tactile inference.
\subsection{Sensorized Actuation Characterization}
\subsubsection{Quasi-static Actuation Test}
\begin{figure}[h!]
    \centering
    \includegraphics[width=0.75\linewidth]{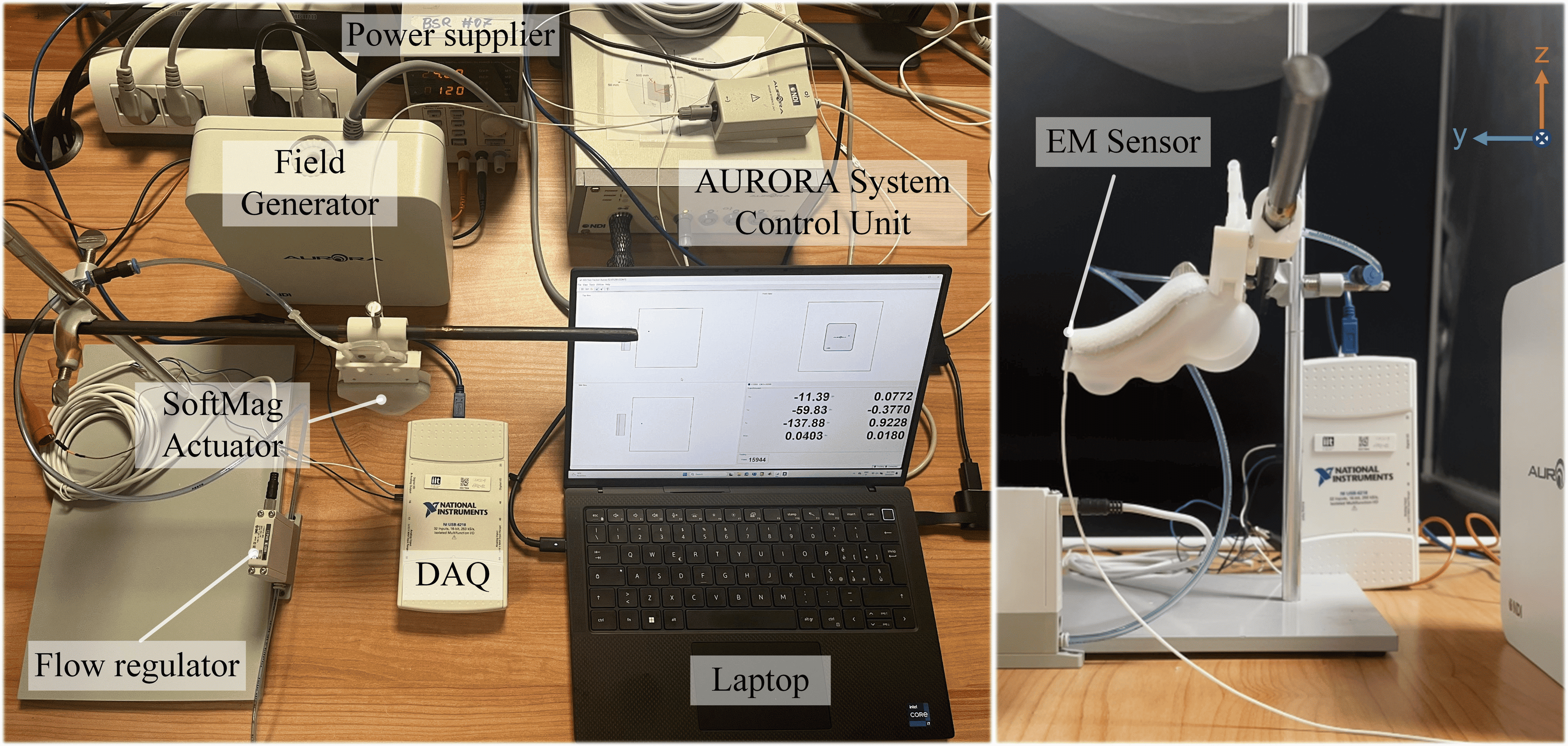}
    \caption{Experimental setup for the quasi-static free-actuation test (left) and the SoftMag actuator in the 35kPa-actuated state with the EM sensor attached (right).}
    \label{fig:6.SetupFree}
\end{figure}
\begin{figure}[h!]
    \centering
    \includegraphics[width=0.75\linewidth]{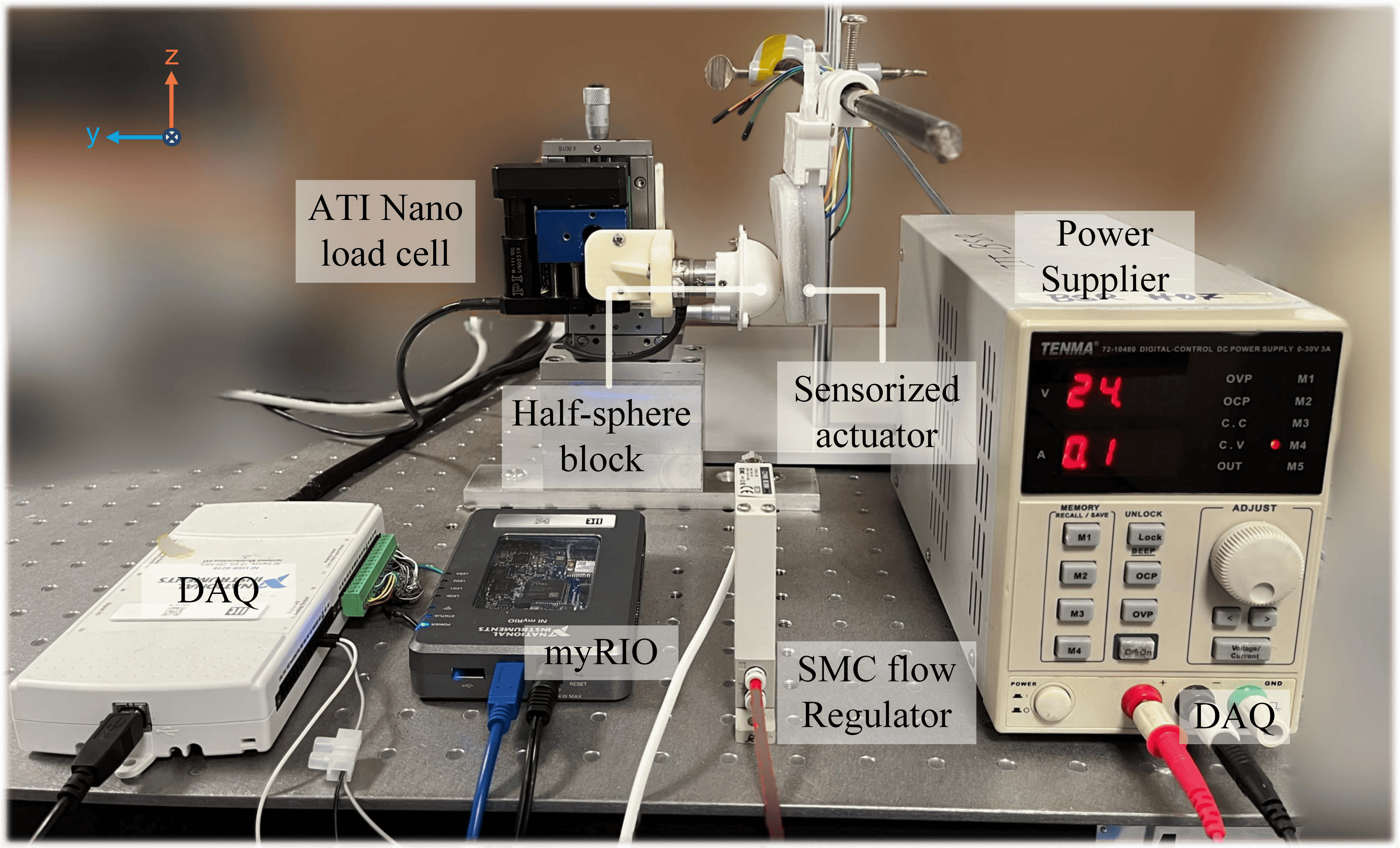}
    \caption{Experimental setup for the quasi-static blocking actuation test.}
    \label{fig:7.SetupBlock}
\end{figure}
\begin{figure}[h!]
    \centering
    \includegraphics[width=0.75\linewidth]{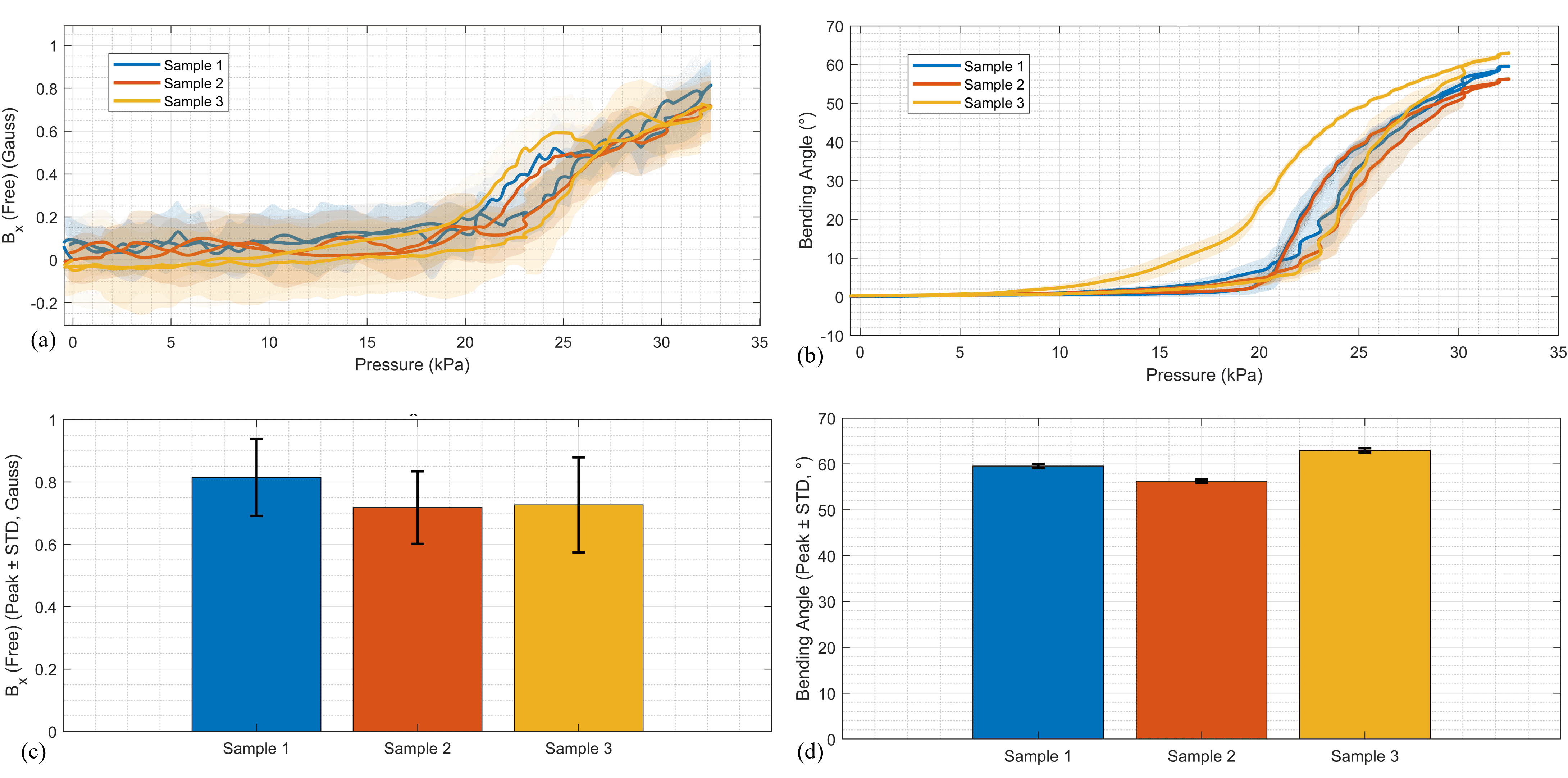}
    \caption{Mean ± standard deviation plots of: (a) $B_x$ and (b) bending angle against actuation pressure across samples in the quasi-static free-actuation test; Comparison of the peak statistics across samples in the quasi-static free-actuation test: (c) $B_x$, and (d) bending angle.}
    \label{fig:8.QuasiResult}
\end{figure}

\noindent\hspace*{1em}Both quasi-static free actuation and quasi-static blocking actuation tests were conducted on three sensorized prototypes to investigate the quasi-static response of the sensing signal and blocking force. Free-actuation tests involved pressurizing the actuator without external constraint, while block-actuation tests were performed against a fixed half-sphere probe to simulate constrained deformation and evaluate force output.  The corresponding setups are illustrated in Figure~\ref{fig:6.SetupFree} and Figure~\ref{fig:7.SetupBlock}, showing the blocked actuation configuration and the free-actuation state with the EM tracking system in place, respectively. Additional details regarding instrumentation and control protocols for both setups are provided in Appendix D.\\
\noindent\hspace*{1em}As shown in Figure~\ref{fig:7.SetupBlock}, the actuator was set to bend only in the YZ plane of the field generator, thus X-axis displacement was assumed negligible and excluded from analysis. Magnetic flux data were processed using the MATLAB pipeline described in Section 3.4.1, while bending angles were computed from tracked tip displacement using Equation (A.1, Appendices A). Figure~\ref{fig:8.QuasiResult} (a–b) shows the mean ± standard deviation of $B_x$ and bending angle across ten actuation cycles per sample. As it is shown, the $B_x$ component exhibited smooth, monotonic growth with increasing pressure and high consistency across cycles, revealing its role as the most reliable axis for deformation sensing. The bending angle followed a similarly consistent trend, confirming the actuator's repeatable kinematic behavior. Peak statistics of $B_x$ and bending angle (Figure~\ref{fig:8.QuasiResult} (c)-(d)) further confirm signal stability and actuation repeatability. Notably, the bending angles achieved here are comparable to (or even exceed) those reported in prior characterizations of the standalone M-PAM actuator\cite{bernabei2023development}, indicating that the integration of the SoftMag sensing layer does not compromise kinematic performance. Moreover, the trends closely align with those observed in the simulation results of Section 3.2, supporting the validity of both the physical and simulated models. Importantly, the consistent flux response across cycles despite no external contact highlights the presence of mechanical parasitic effects, where internal deformation alone induces measurable magnetic signal changes. This aligns with predictions from the simulation (Section 3.2) and emphasizes the importance of compensating for such actuation-induced signals in tactile soft robotic systems, as later discussed in Section 5.
\begin{figure}[h!]
    \centering
    \includegraphics[width=0.75\linewidth]{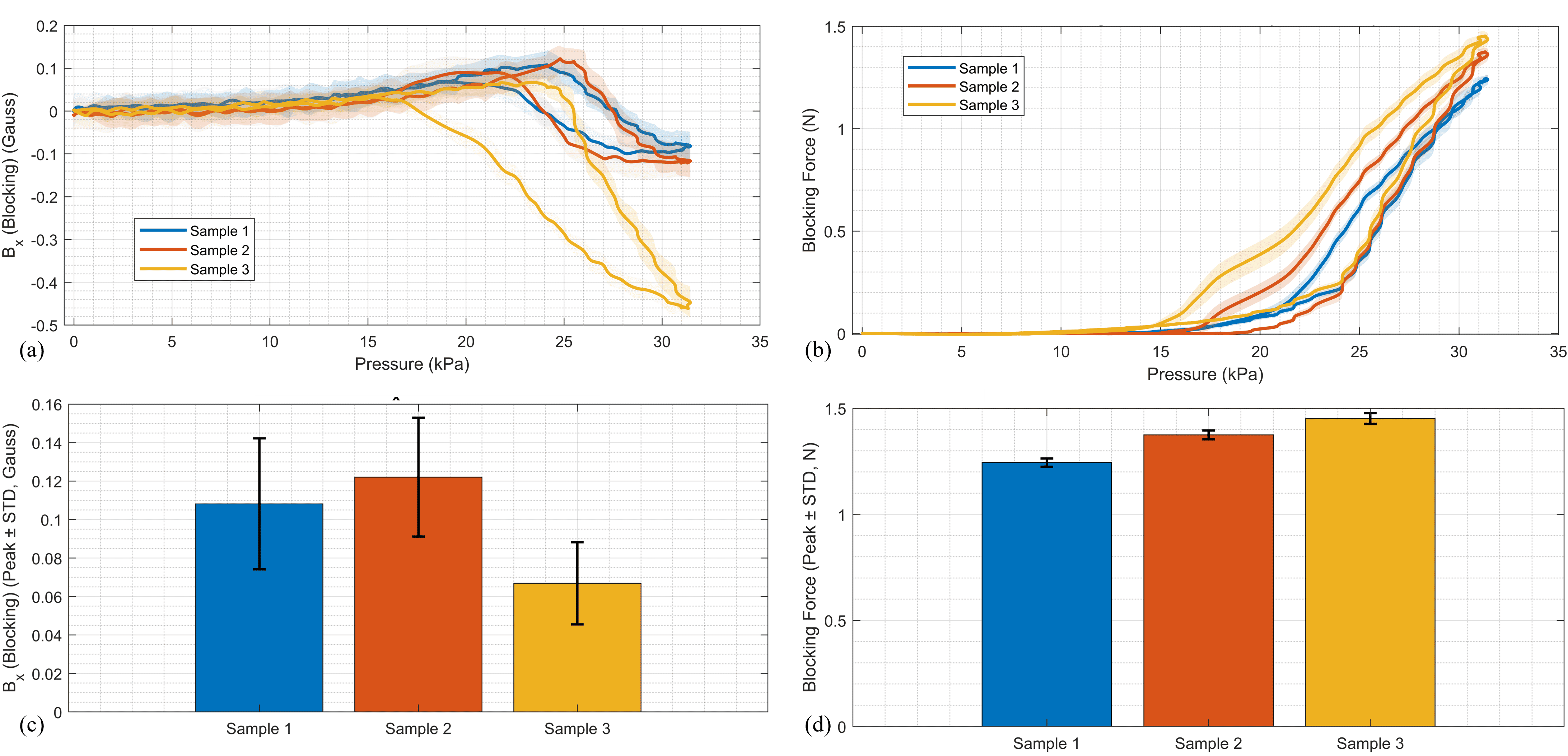}
    \caption{Mean ± standard deviation plots of: (a) $B_x$ and (d) blocking force against actuation pressure across samples in the quasi-static blocking-actuation test; Comparison of the peak statistics across samples in the quasi-static blocking-actuation test: (c) $B_x$ and (d) blocking force.}
    \label{fig:9.BlockResult}
\end{figure}

\noindent\hspace*{1em}The quasi-static blocking-actuation tests were conducted using the setup depicted in Figure~\ref{fig:6.SetupFree} to assess the SoftMag actuator’s sensing response and force output under constrained conditions. Magnetic data were processed using the MATLAB pipeline described in Section 3.4.1, while blocking force readings were directly extracted and analyzed. Figure~\ref{fig:9.BlockResult} (a–b) presents the mean ± standard deviation of the $B_x$ component and blocking force across ten actuation cycles per sample. While Bx initially increases with pressure, it exhibits nonlinear behavior at higher pressures, likely due to interactions between actuator deformation and probe geometry. In contrast, the blocking force grows steadily and consistently across samples, reaching ~1.4N at 35kPa. The peak statistics in Figure~\ref{fig:9.BlockResult} (c–d) confirm this trend: force output remains tightly grouped, while $B_x$ shows more variability due to contact-induced alignment differences and structural asymmetries.\\
\noindent\hspace*{1em}These results demonstrate the actuator’s reliable mechanical output and further illustrate how mechanical parasitic effects can influence sensor readings in contact-rich scenarios. This highlights the need for robust decoupling strategies, as addressed in Section 5.

\subsubsection{Step Actuation Test}
\begin{figure}[h!]
    \centering
    \includegraphics[width=0.75\linewidth]{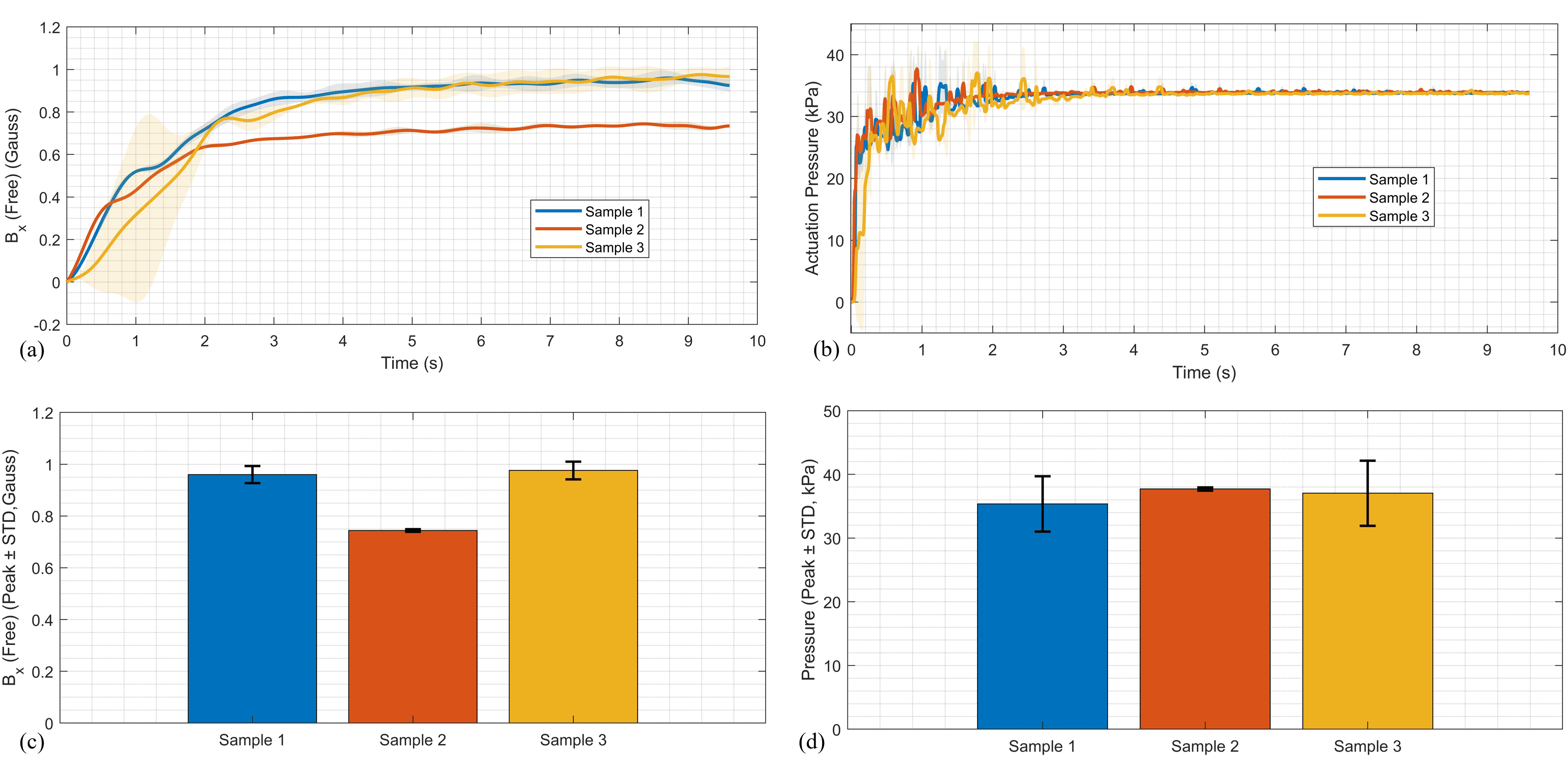}
    \caption{Mean ± standard deviation plots of: (a) $B_x$ and (b) real-time actuation pressure across samples in the step free-actuation test; Comparison of the peak statistics across samples in the step free-actuation test: (c) $B_x$ and (d) real-time actuation pressure.}
    \label{fig:10.StepResult}
\end{figure}

\noindent\hspace*{1em}The SoftMag actuator’s dynamic behavior was tested by inflating with a step input pressure the actuator, being the latter in both free and blocked conditions. In the free-actuation configuration, each prototype was rapidly inflated to 35kPa and held for 1s before deflation, with magnetic and pressure data recorded at 50Hz. Each of the three samples underwent two repetitions. Only the inflation phase was analyzed to focus on the rising response. Magnetic flux and kinematic data were processed following the same protocol described in Section 3.4.1.\\
\noindent\hspace*{1em}Figure~\ref{fig:10.StepResult} (a–b) shows the average $B_x$ response and actuation pressure over time, with shaded envelopes indicating trial-to-trial variation. Among all flux components, $B_x$ exhibited the most consistent and monotonic growth, demonstrating stable sensitivity to deformation. The observed differences among samples (especially in the early phase for Sample 3) likely reflect variations in chamber compliance or material damping. Pressure curves confirmed rapid rise to the target value within ~1.5s and remained stable throughout inflation. The peak values for $B_x$ and pressure across samples are summarized in Figure~\ref{fig:10.StepResult} (c–d). The $B_x$ component peaks highest in Samples 1 and 3 with low standard deviation. The pressure peaks were consistently close to 35kPa, validating the uniformity of input conditions. These results highlight the SoftMag actuator’s dynamic sensing repeatability in unconstrained actuation and underscore the prominence of $B_x$ as a robust indicator of actuation response.

\begin{figure}[h!]
    \centering
    \includegraphics[width=0.75\linewidth]{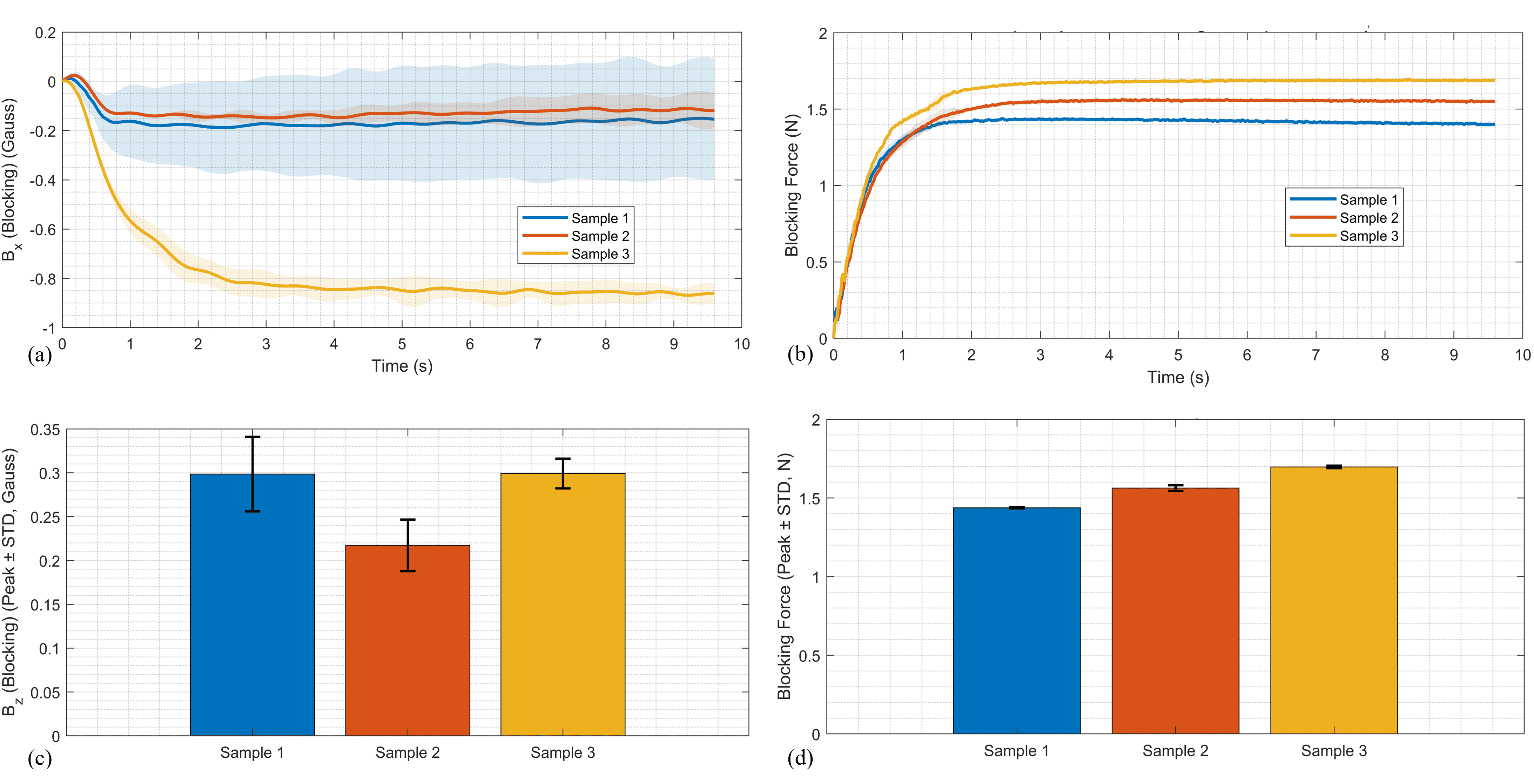}
    \caption{Mean ± standard deviation plots of: (a) $B_x$ and (b) blocking force across samples in the blocking-actuation test; Comparison of the peak statistics across samples in the blocking-actuation test: (c) $B_x$ and (d) blocking force.}
    \label{fig:11.BlockResult}
\end{figure}

\noindent\hspace*{1em}The step blocking-actuation test was conducted to evaluate the dynamic behavior of the SoftMag actuator and its integrated magnetic sensing under constrained inflation. The inflation, preprocessing, including interpolation and alignment, followed the same protocol as in the free-actuation test (Section 3.4.1). Figure~\ref{fig:11.BlockResult} (a–b) presents the mean ± standard deviation of $B_x$ and blocking force overtime across all three samples. Under blocking, all samples exhibited a rapid negative deflection in $B_x$ during actuation. The response stabilized quickly after ~2s and was maintained throughout the inflation phase. The blocking force rose consistently in all samples, saturating around 1.5–1.8N. These results confirm the actuator’s ability to produce steady contact forces and highlight $B_x$’s sensitivity to constrained deformation. Figure~\ref{fig:11.BlockResult} (c–d) compares the peak values of $B_x$ and blocking force across samples. While $B_x$ showed greater inter-sample variability due to contact misalignment or probe positioning, the force output remained tightly clustered, confirming reliable mechanical performance under blocked conditions.\\
\noindent\hspace*{1em}To assess the actuator’s temporal behavior under step actuation, standard dynamic metrics were extracted from both pressure and magnetic flux signals (definitions in Appendix D). Figure~\ref{fig:12.StepofS2} illustrates the annotated mean response curves for Sample 2 as the representative among the three tested samples. In the free-actuation condition (Figure~\ref{fig:12.StepofS2} (a) and (c)), $B_x$ smooth, monotonic growth aligned with the pressure rise, while By responded more gradually. $B_z$ rose quickly but plateaued or fluctuated, making it less reliable for steady-state tracking. The pressure signal reached steady state within $~1.5s$. In the blocking-actuation condition (Figure~\ref{fig:12.StepofS2} (b), (d)), all flux components showed an initial negative deflection, with Bx stabilizing clearly, whereas $B_y$ and $B_z$ were more erratic, likely due to torsional or compressive effects near the sensor. The blocking force rose rapidly and stabilized within $~2–3s$. These results demonstrate consistent dynamic performance across modalities, with Bx and By offering the most stable sensing behavior. Full comparisons and metrics for all samples are available in Appendix D. The findings highlight the importance of axis-specific calibration and support the use of decentralized control strategies in multi-actuator systems.

\begin{figure}[h!]
    \centering
    \includegraphics[width=0.75\linewidth]{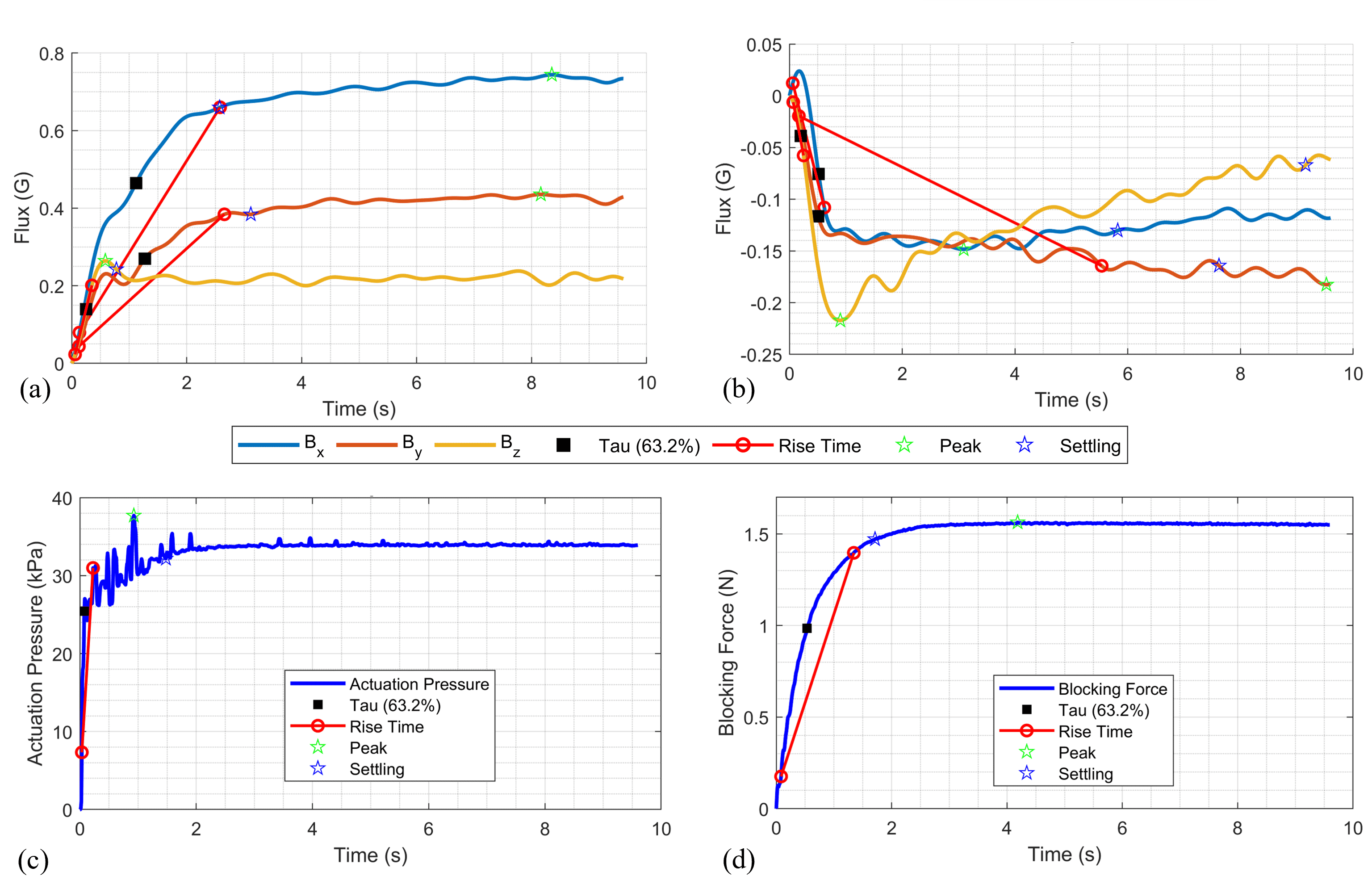}
    \caption{Dynamic analysis of the average responses of Sample 2 in the step actuation test: (a) Flux responses analysis in free-actuation; (b) Flux responses analysis in blocking-actuation; (c) Real-time actuation pressure analysis; (d) Blocking force analysis.}
    \label{fig:12.StepofS2}
\end{figure}

\subsection{Fatigue Test}
\noindent\hspace*{1em}Fatigue tests were conducted to evaluate the durability of the SoftMag actuator under repeated inflation-deflation cycles. Each sample was cyclically actuated from 0 to 35kPa over 15s intervals at 50Hz, using the same setup as in the quasi-static free actuation test (Section 3.5). Each actuator was subjected to continuous cyclic actuation until mechanical failure occurred. Failure here was defined as rupture or delamination of the silicone structure, resulting in an inability to maintain internal pressure. No sensor drift or misalignment occurred prior to failure, indicating stable sensor integration. The total number of cycles completed by each actuator (including those accumulated during prior sensorized actuation experiments) was recorded as a metric of fatigue endurance. The failure occurred at 531, 409 and 383 cycles for Samples 1, 2, and 3, respectively.  This outcome suggests that subtle microstructural differences such as foam porosity, magnet alignment, bonding uniformity, and wall thickness introduced during fabrication may significantly influence fatigue performance. While the observed endurance is reasonable for soft actuators at moderate pressures, future improvements will be key to enabling long-term deployment in repetitive tasks.

\section{Towards the SoftMag Gripper} 
\subsection{System Framework} 
\begin{figure}[h!]
    \centering
    \includegraphics[width=0.75\linewidth]{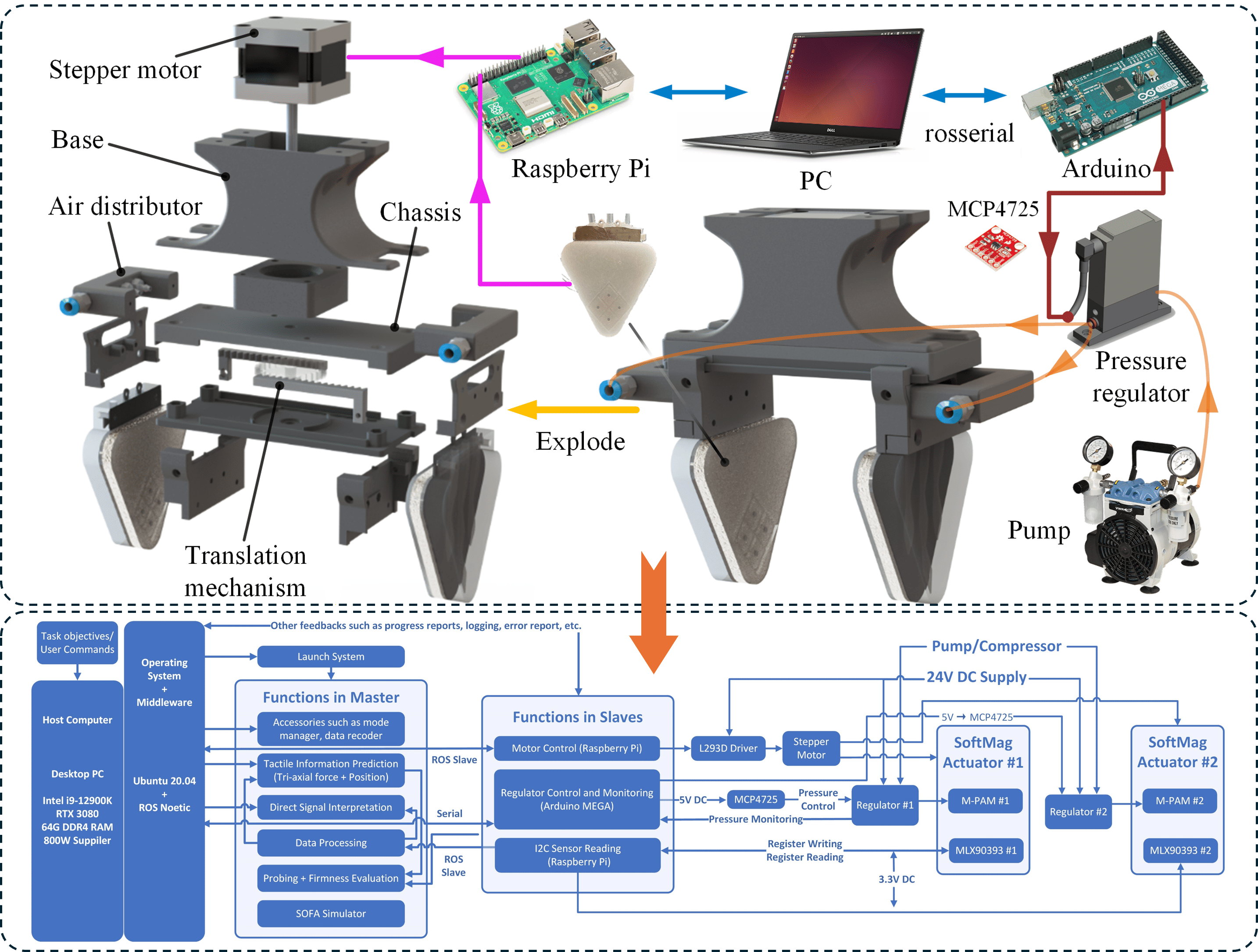}
    \caption{Mechatronic composition of the two-finger gripper platform (top) and schematics of its system architecture (bottom).}
    \label{fig:13.ToFPlatform}
\end{figure}

\noindent\hspace{1em}After the characterization data was assessed, the SoftMag actuators were further extended into a compact, modular, and sensorized two-finger system capable of adaptive grasping and real-time tactile feedback. As shown in Figure~\ref{fig:13.ToFPlatform}, each actuator is mounted on a rack that translates via a gear-driven mechanism powered by a stepper motor, enabling symmetric finger movement. Individual pneumatic control is provided to each actuator through a dual-regulator configuration, allowing synchronized deformation even in the presence of minor mechanical differences. An internal air distributor routes pressure through three channels to match the actuator’s trident airway design. The complete system operates through a distributed ROS-based architecture. A high-performance PC acts as the master, managing control logic, sensor processing, and real-time tactile inference. It communicates with two ROS-enabled devices: a Raspberry Pi for motor control and sensor reading, and an Arduino Mega for pressure regulation and feedback. Actuation is achieved using DAC-driven regulators, while sensor data and force predictions are visualized and managed through a custom graphical interface. Details about the mechatronic components can be found in Appendix E. A real-time system demonstration is available in Video 2.
\subsection{Payload and General Grasping Test} 
\begin{figure}[h!]
    \centering
    \includegraphics[width=0.75\linewidth]{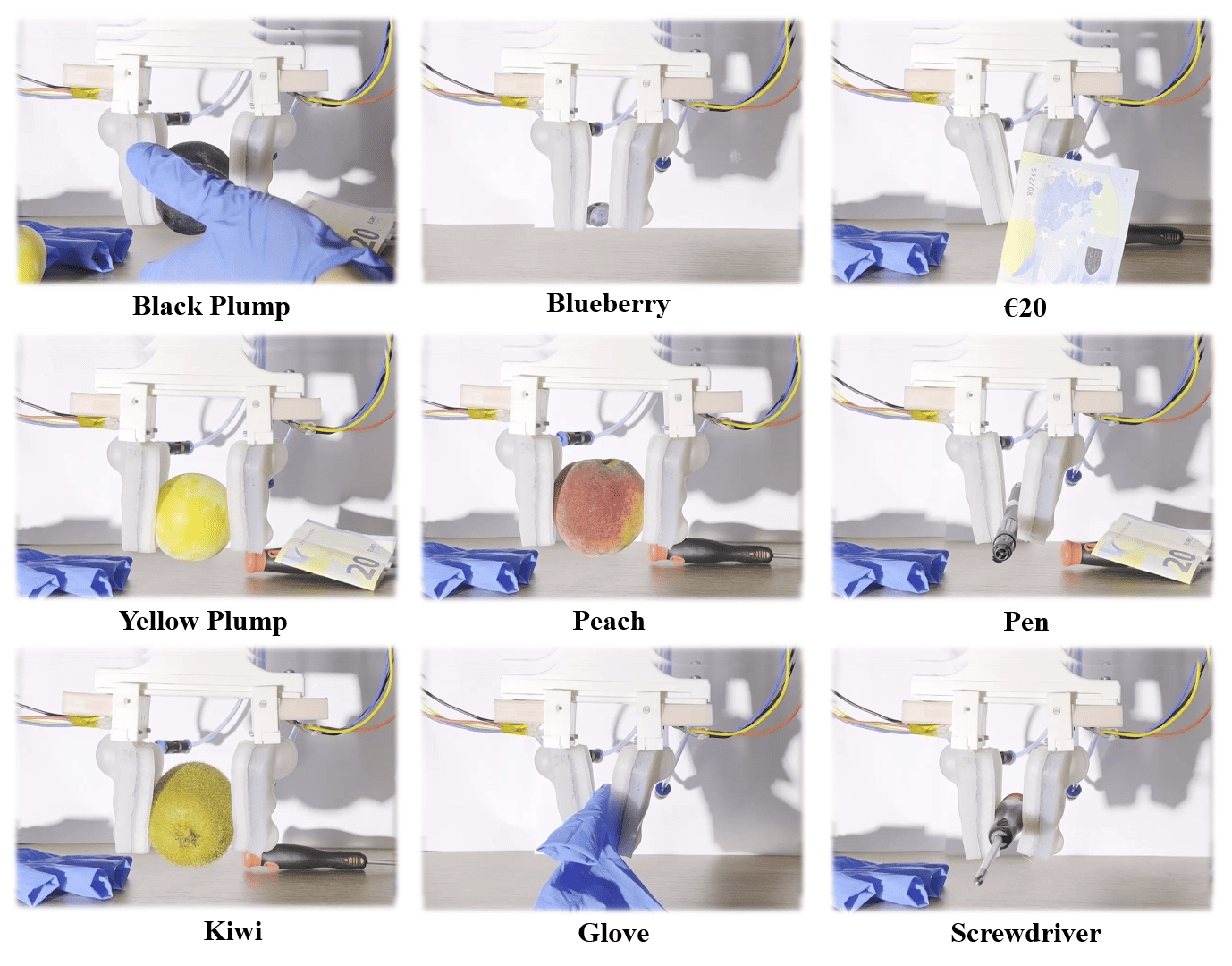}
    \caption{Demonstration of the SoftMag gripper performing grasping tasks on various objects.}
    \label{fig:14.Demo}
\end{figure}

\noindent\hspace*{1em}To assess the SoftMag gripper’s load-bearing capability, a custom payload test was conducted using a dual-actuator setup with a gradually increasing load. As detailed in Appendix F, the procedure involved transferring weight from a scale to the gripper and incrementally adding mass until slippage was observed. The gripper successfully supported a maximum payload of up to 833.8g, yielding a payload-to-weight ratio of ~8.9:1. This demonstrates the gripper’s ability to handle a broad range of produce, from light fruits like apricots (~40g) to heavier ones like coconuts (~700g), supporting its applicability in agricultural tasks such as picking, sorting, and packing (University of Georgia Extension, 2025; Weight of Things, 2025).\\
\noindent\hspace*{1em}Beyond payload testing, the system also demonstrated strong adaptability in general grasping tasks. As shown in Figure~\ref{fig:14.Demo}, the gripper effectively handled various objects with different shapes, sizes, and textures, validating its practical utility in diverse scenarios. The payload and grasping tests can also be found in Video 3.

\subsection{Magnetic Interference Evaluation with Ferrite-Coated Objects}
\begin{figure}[h!]
    \centering
    \includegraphics[width=0.75\linewidth]{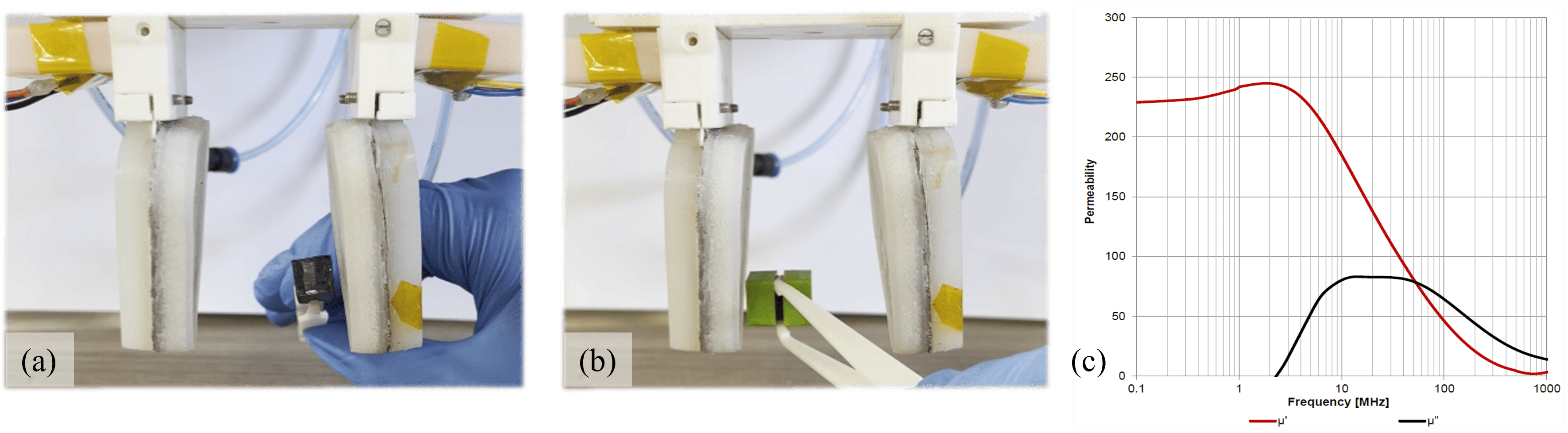}
    \caption{Magnetic interference test using ferrite-skinned tube combined with cap shells; (a) Approaching test using ferrite-skinned tube; (b) Approaching test using ferrite-skinned tube with 3.5 mm cap shell; (c) Permeability vs. frequency plot of the tested FSFS sheet.}
    \label{fig:15.MagInterference}
\end{figure}

\noindent\hspace*{1em}To assess magnetic interference from ferrite materials during proximity or grasping tasks, controlled approach tests were conducted using a ferrite-skinned object. The goal was to identify a safe operational distance to prevent magnetic distortion in the SoftMag sensing system. The test object was a solid ABS cube coated with a flexible FSFS (Flexible Sintered Ferrite Sheet, part number 354003, Würth Elektronik GmbH \& Co. KG, Germany). The object was brought near the actuator in its unactuated state at different separation distances (0, 3.5, 5.0, and 7.5\,mm), simulating real-world approach scenarios. A plastic tweezer ensured no additional magnetic disturbance. Figure~\ref{fig:15.MagInterference}(a–b) illustrates representative approach trials, and Video~4 shows the entire test procedure.\\
\noindent\hspace*{1em}Results revealed that at direct contact (0\,mm), all three magnetic axes exhibited noticeable deviations (up to 0.4\,G). At 3.5\,mm, variations dropped below 0.1\,G. Disturbances at 5.0\,mm were minimal, and at 7.5\,mm, signal shifts were indistinguishable from baseline noise. These findings define 7.5\,mm as a conservative minimum safety buffer to avoid false tactile readings from ferrite interference. Compared to the tested ferrite sheet ($\mu' \approx 230$), metals like soft iron and steel have significantly higher relative permeabilities ($\mu_r > 1{,}000$), likely causing more severe interference. Therefore, we recommend avoiding direct grasping of high-permeability objects ($\mu_r \approx 100$) and maintaining at least 10\,mm of clearance when interacting with unknown or potentially magnetic materials. This guideline ensures reliable tactile sensing, particularly in unstructured environments.
\subsection{Tactile Inference Using Multi-Task Network}
\begin{figure}[h!]
    \centering
    \includegraphics[width=0.75\linewidth]{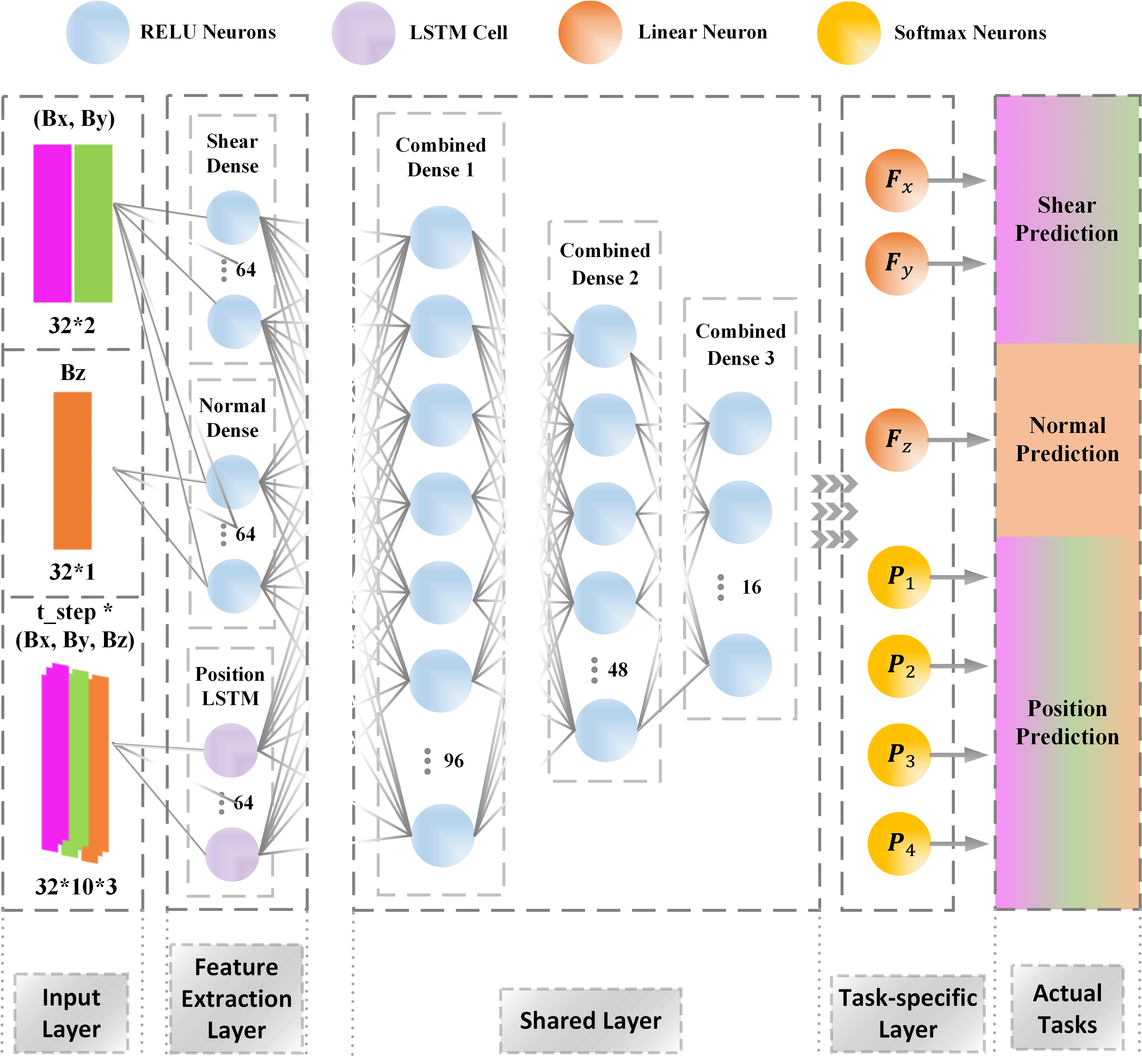}
    \caption{Schematic of the multi-task learning model for real-time tactile inference.}
    \label{fig:16.Learning}
\end{figure}
\noindent\hspace*{1em}The SoftMag system leverages a multi-task neural network to predict tactile information, including shear force, normal force, and contact position, directly from magnetic flux signals. As shown in Figure~\ref{fig:16.Learning}, the model consists of a shared input that branches into task-specific pathways. Dense layers with ReLU activation are used for force estimation, while a Long Short-Term Memory (LSTM) layer with Tanh activation captures time-dependent patterns critical for position inference, as previously suggested by slope analysis of flux signals over time. A Softmax layer is used for position classification output. Magnetic flux inputs are normalized to the range $(-1, 1)$ and force values to the range $(0, 1)$ using a MinMaxScaler. The dataset is split into $70\%$ training, $15\%$ validation, and $15\%$ test sets. Training is conducted for a maximum of 200 epochs with a batch size of 32, using the Adam optimizer with gradient clipping (parameters as in the default Keras implementation). Early stopping with a patience of 10 epochs is applied based on validation loss to prevent overfitting. The network is trained using a combined loss function:
\begin{equation}
    L_{\text{total}} = \lambda_1 \, \mathrm{MSE}(\hat{y}_{\text{shear}}, y_{\text{shear}}^{\text{true}})
    + \lambda_2 \, \mathrm{MSE}(\hat{y}_{\text{normal}}, y_{\text{normal}}^{\text{true}})
    + \lambda_3 \, \mathrm{CrossEntropy}(\hat{y}_{\text{pos}}, y_{\text{pos}}^{\text{true}})
\tag{4.1}
\end{equation}
\noindent\hspace*{1em}Performance on the test set after 200 epochs shows a total loss of 0.098, with shear and normal losses of 0.014 and 0.033, respectively, and position loss of 0.091. The mean absolute errors (MAE) are 0.085 (shear) and 0.138 (normal), and the position classification accuracy reaches 0.960. This architecture balances task-specific feature extraction with shared representation learning, enabling efficient and simultaneous inference of all tactile parameters in real time. Its performance and inference behavior are illustrated in Video 1.

\section{Mechanical Parasitic Effect and Decoupling Strategy} 
\begin{figure}[h!]
    \centering
    \includegraphics[width=0.7\linewidth]{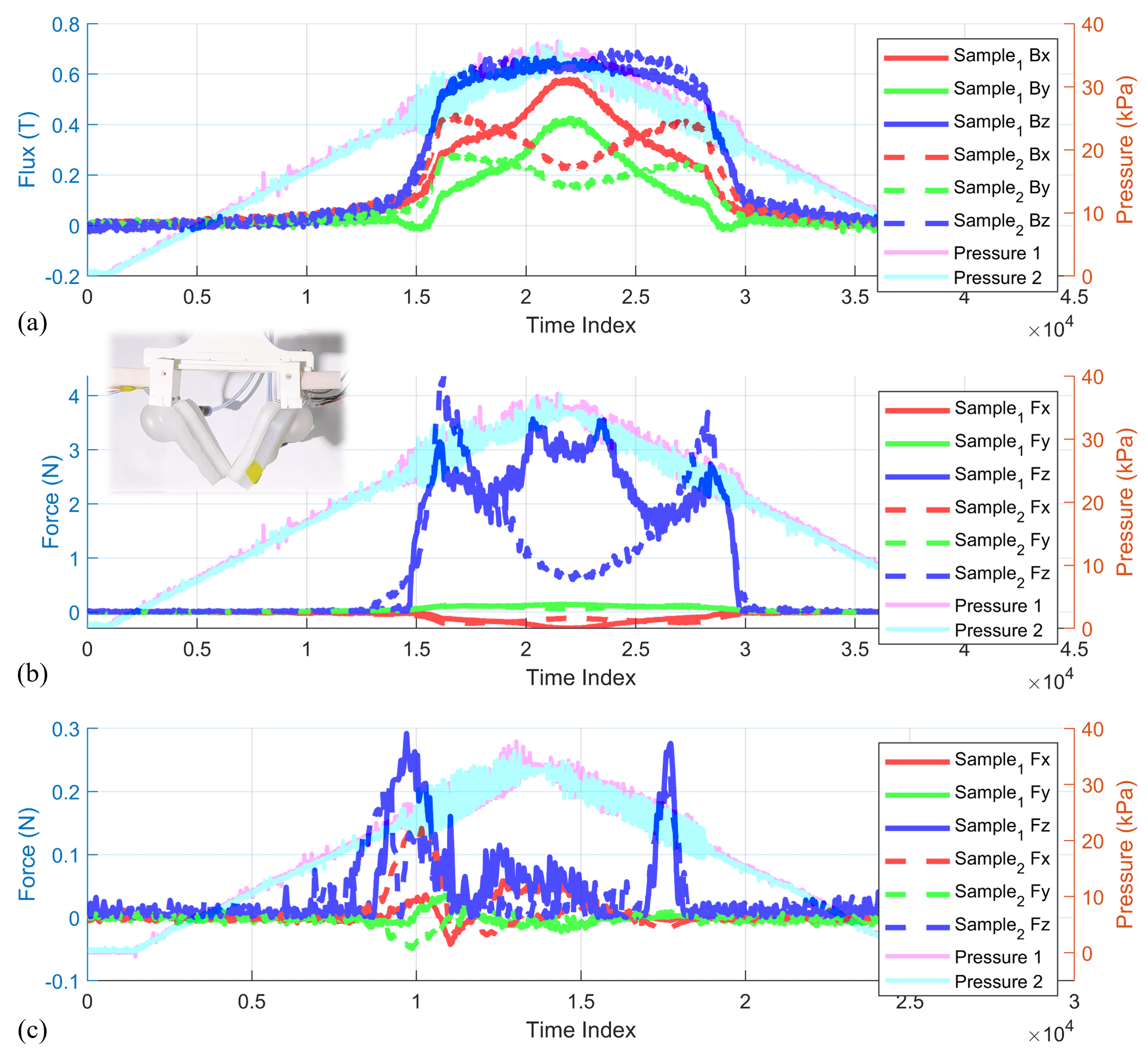}
    \caption{Parasitic decoupling performance under ramp actuation: S1 and S2 represent the sensor readings from the SoftMag actuators 1 and 2; Pressure 1 and 2 represent the pressure sensor readings from the SoftMag actuators 1 and actuator 2.}
    \label{fig:17.Parasitic}
\end{figure}
\noindent\hspace*{1em}As presented in Sections~3.2 and~3.5.1, the mechanical coupling between actuation and sensing components in SoftMag actuators can introduce parasitic effects that distort the desired tactile output during pneumatic inflation or deflation. To mitigate this, a Multi-Layer Perceptron (MLP) neural network is introduced to predict and subtract the actuator-induced magnetic flux from raw sensor signals. Each MLP was trained per actuator using a dataset acquired by applying the pressure from 0 to 35\,kPa and back in 0.1\,kPa steps, repeated five times. Input features were filtered pressure readings; targets were the corresponding 3D flux components. The networks used two hidden layers (64 and 32 neurons), ReLU activation, the Adam optimizer, and early stopping (validation split: 0.1, max epochs: 200). Once trained, the model estimates the actuation-induced flux, which is subtracted to yield the decoupled signal:
\begin{equation}
\vec{F}_{\text{decoupled}}, \vec{\mathcal{P}}_{\text{decoupled}} = \mathrm{MultiTask}\left( \vec{B} - \mathrm{MLP}_1^3(P) \right)
\tag{5.1}
\end{equation}
\noindent\hspace*{1em}Here P represents the actuation pressure. During real-time operation, the final decoupled flux signal is obtained by subtracting the predicted actuation-induced flux from the raw measurements. This approach effectively mitigates the parasitic effect, improving the accuracy and reliability of subsequent force and position estimations.\\
\begin{figure}[h!]
    \centering
    \includegraphics[width=0.75\linewidth]{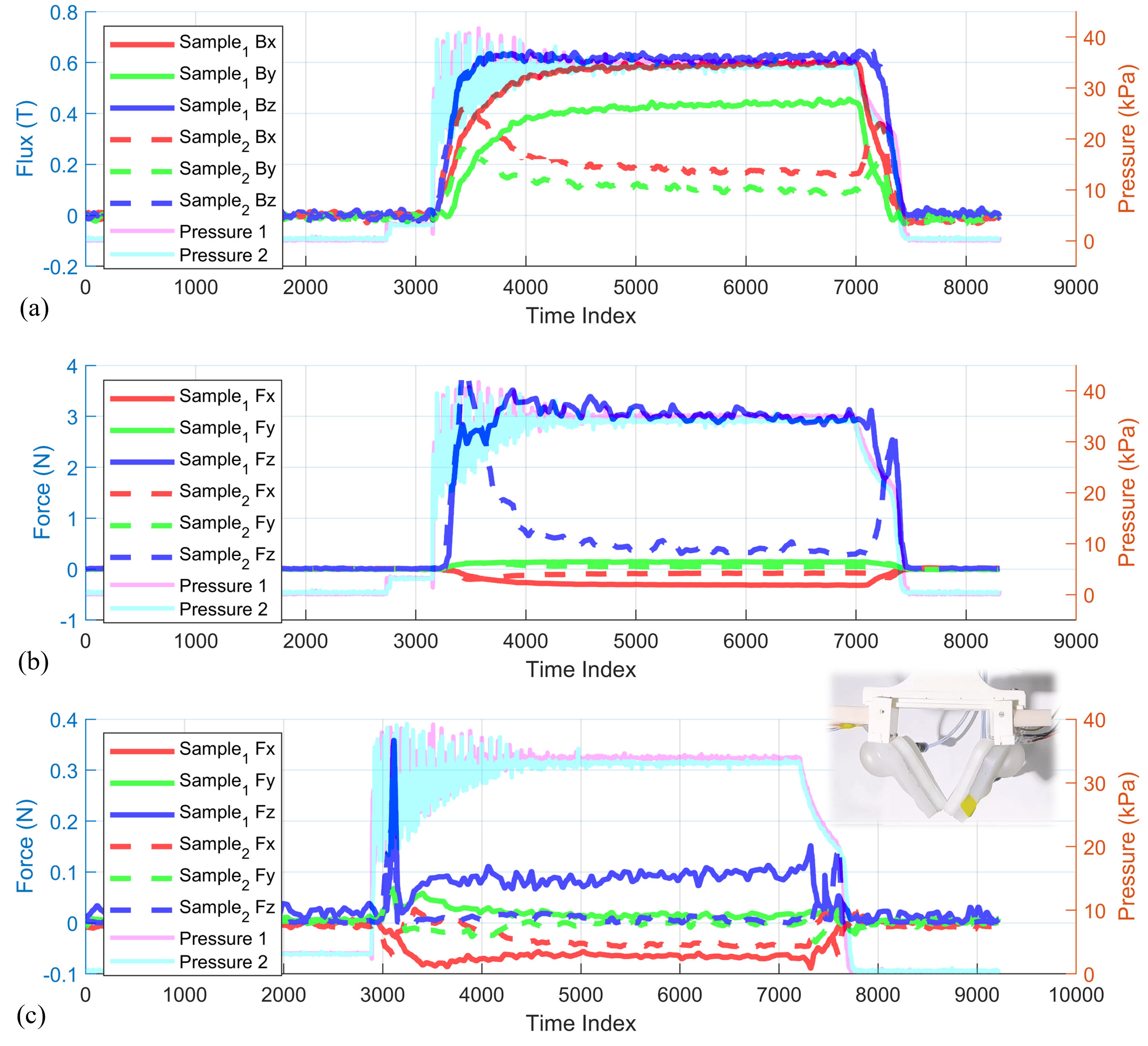}
    \caption{Decoupling performance under step actuation.}
    \label{fig:18.StepPerf}
\end{figure}\\
\noindent\hspace*{1em}To validate the effectiveness of the proposed decoupling method, comparative tests were conducted using both ramp and step actuation profiles. Figure~\ref{fig:17.Parasitic} shows results under a ramp actuation profile. It can be observed that without decoupling, the flux signals from S1 and S2 drift in tandem with pressure increases, producing inflated force estimates (Figure~\ref{fig:17.Parasitic} (a)-(b)). With decoupling enabled, force signals are greatly reduced ($<0.3N$), confirming suppression of actuation-induced artifacts (Figure~\ref{fig:17.Parasitic} (c)). Figure~\ref{fig:18.StepPerf} presents results for a step actuation input. Again, raw signals show sharp transients aligned with pressure spikes, distorting tactile inference (Figure~\ref{fig:18.StepPerf} (a)-(b)). Once decoupling is applied, the distortions are considerably mitigated, with a marked reduction in both baseline drift ($<0.4N$ step response) and high-frequency noise (Figure~\ref{fig:18.StepPerf} (c)). While residual artifacts are still observed during the inflate and release phases, these results demonstrate that the neural network-based decoupling approach substantially improves sensor signal quality. Video 5 shows both the concept of the mechanical parasitic effect and the above validation process.\\
\begin{figure}[h!]
    \centering
    \includegraphics[width=0.75\linewidth]{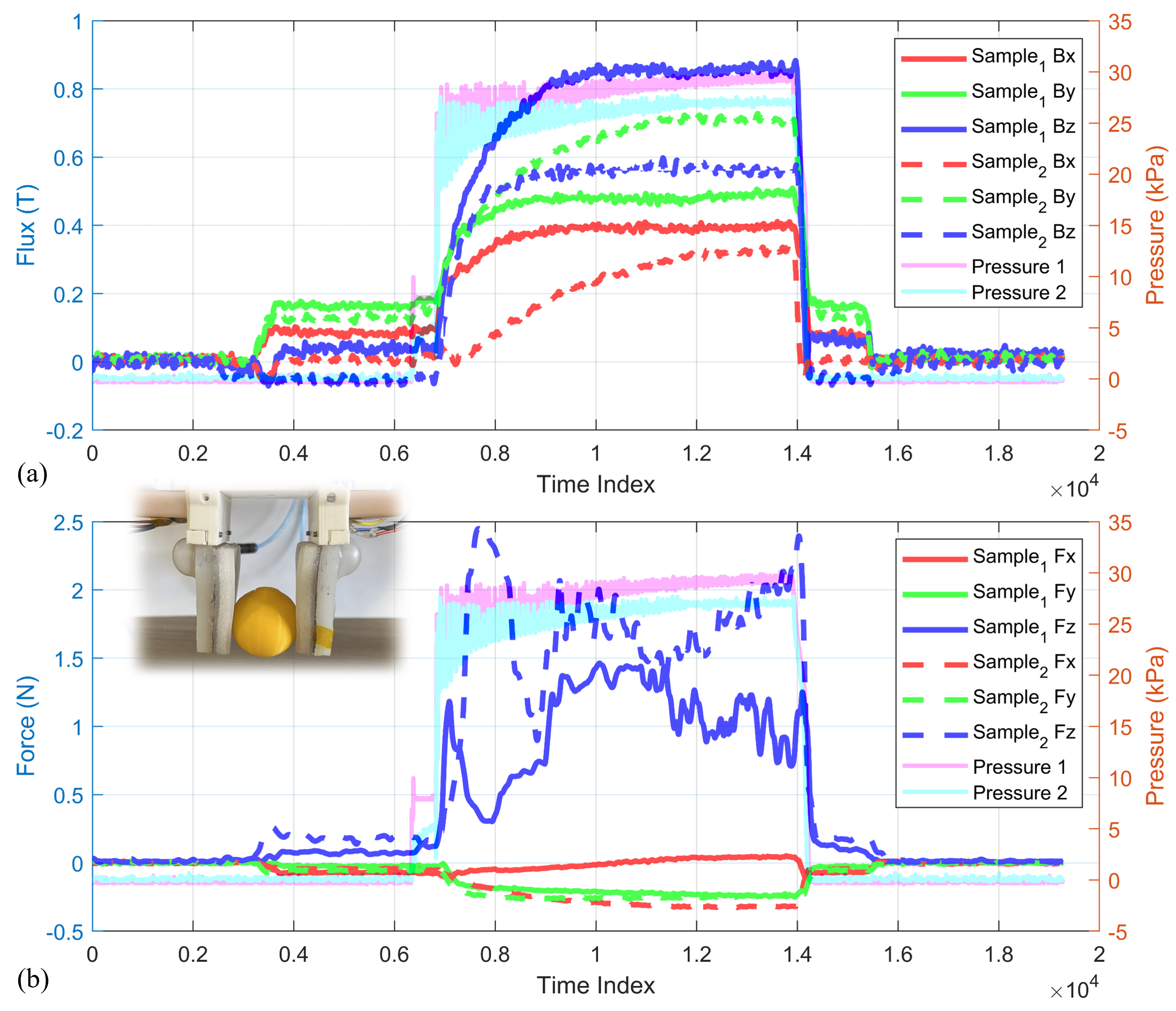}
    \caption{Decoupling performance during grasp-and-release cycle.}
    \label{fig:19.Grasp}
\end{figure}
\noindent\hspace{1em}The method was further evaluated in a realistic grasp-and-release task with a step input. As shown in Figure~\ref{fig:19.Grasp} (a), raw flux data exhibit sharp changes due to both actuator deformation and object contact, while decoupled force estimates (Figure~\ref{fig:19.Grasp} (b)) remain stable during grasp hold and release, with minimal transients and good consistency across both actuators. This confirms that the decoupling method preserves meaningful contact signals while eliminating actuator-driven noise, enabling robust tactile sensing in dynamic, in-contact scenarios.

\section{Firmness Evaluation by soft probing}
\noindent\hspace*{1em}Building upon the tactile prediction model and parasitic decoupling strategy introduced before, a firmness estimation function is developed under the SoftMag gripper system. Herein, the firmness $\phi_{obj}$ of an object is quantified in a manner akin to Hooke's law, where many literatures treat fruits as a spring model\cite{liu2023prediction,ahmadi2012dynamic}. The proposed method estimates object firmness based on the gripper's internal force and pressure variations during dynamic probing. \\
\noindent\hspace*{1em}Firmness is formulated as:
\begin{equation}
    \phi_{\text{obj}} = b \cdot \exp\left(
    \frac{a \cdot \left(
    \sup_{t \in T} \Delta F_{\text{probing}}^{(A_1)}(t) + 
    \sup_{t \in T} \Delta F_{\text{probing}}^{(A_2)}(t)
    \right)}{2 \int_0^T \frac{dP(t)}{dt} \, dt}
    \right)
\tag{6.1}
\end{equation}

where $\Delta F_{\text{probing}}^{(A_1)}$ represents the probing force predicted by actuator 1, 
and $\int_0^T \frac{dP(t)}{dt} \, dt$ represents the cumulative rate of pressure change during the probing process. Constants $a$ and $b$ are empirically determined once, based on the probing dataset, to normalize and amplify the firmness metric into a numerically interpretable range. The same fixed values are used across all reported objects and tests, without per-object re-tuning.

\noindent\hspace{1em}The entire probing process is illustrated in Figure~\ref{fig:20.PLAball}. The procedure begins with a 28.5kPa step input to achieve a stable initial grasp. Then, a 4kPa square-wave pressure modulation is applied periodically, each cycle lasting 5 seconds. The resulting force response is sampled after 2 seconds of pressurization to avoid transient disturbances and to capture steady deformation effects. This approach mimics human finger tapping when assessing fruit ripeness. Video~6 provides a real-time demonstration of the proposed firmness estimation method.

\subsection{Firmness Estimation Test}
\begin{figure}[h!]
    \centering
    \includegraphics[width=0.75\linewidth]{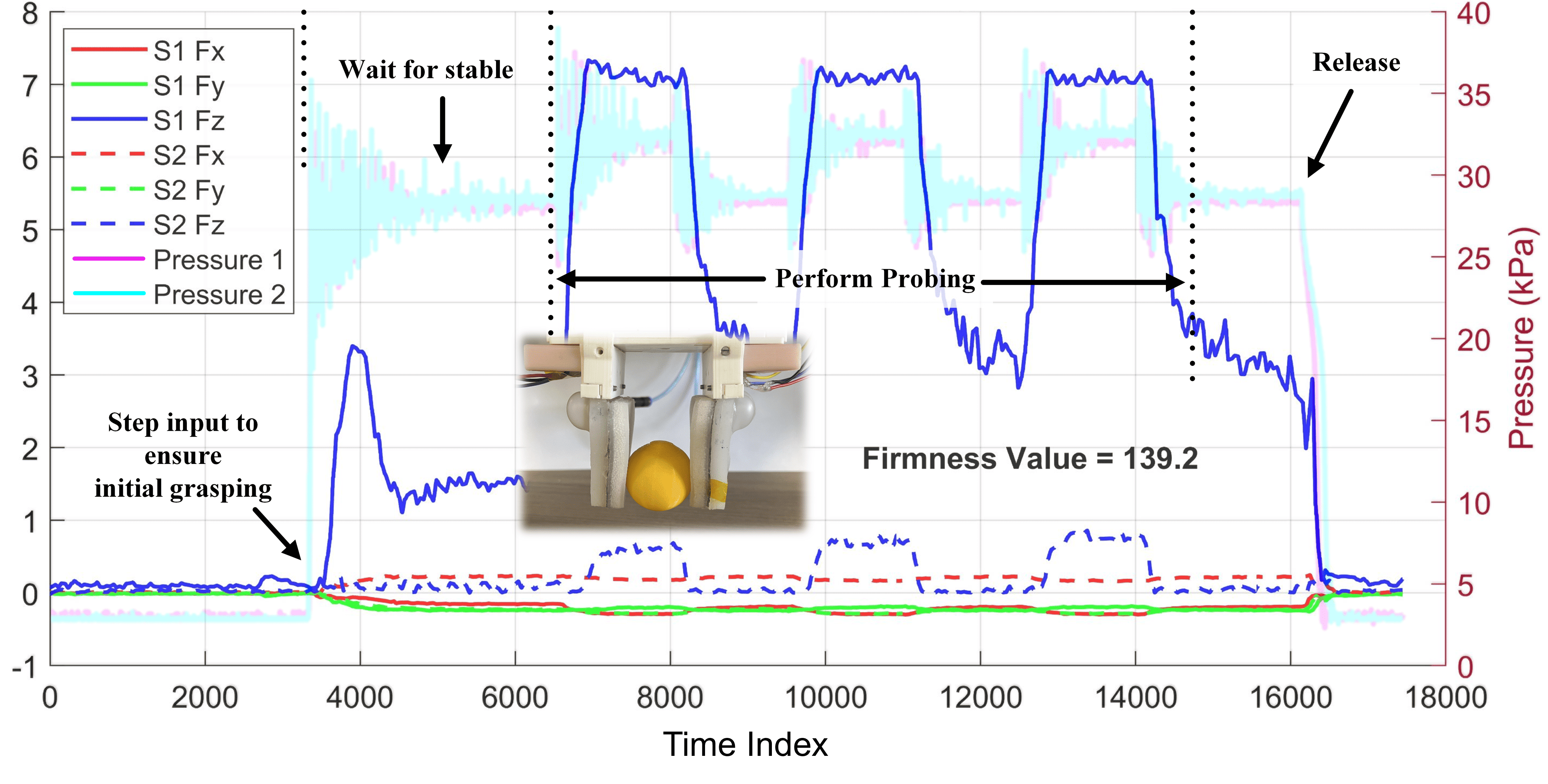}
    \caption{Force vs. pressure plot during probing of a rigid PLA ball.}
    \label{fig:20.PLAball}
\end{figure}

\begin{figure}[h!]
    \centering
    \includegraphics[width=0.6\linewidth]{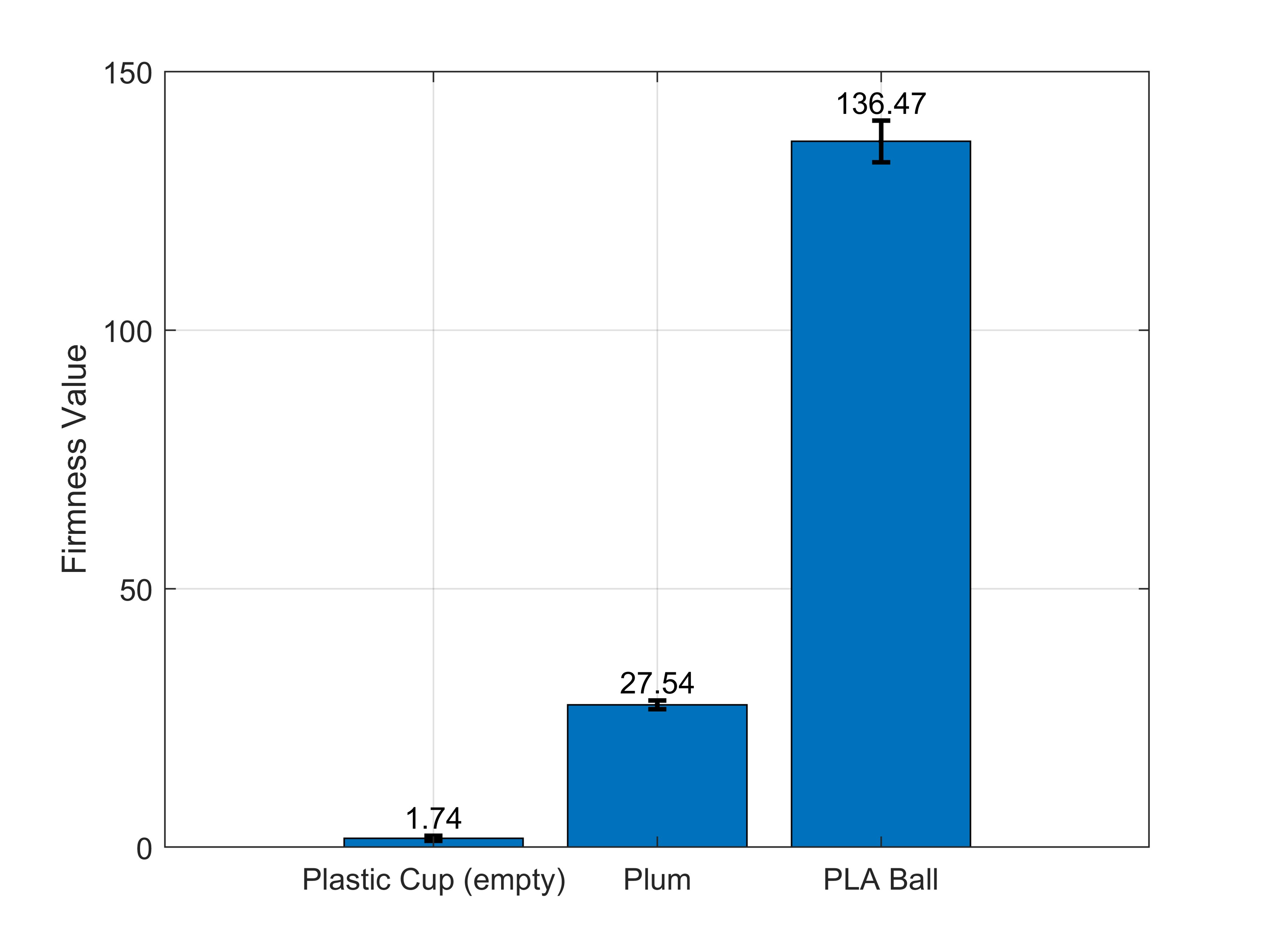}
    \caption{Average firmness estimation of the three tested objects.}
    \label{fig:21.Firmness}
\end{figure}

\noindent\hspace{1em}To validate this strategy, three objects with varying stiffness levels were tested: a rigid polylactic acid (PLA) ball, a medium-soft fresh plum, and a very soft empty disposable plastic cup. For each case, the predicted force signals from the SoftMag gripper were analyzed during the probing cycles. The PLA ball, representing a rigid object, exhibited clear and strong force peaks during each pressure increment, as shown in Figure~\ref{fig:20.PLAball} In contrast, the soft plastic cup showed minimal force response with small, damped variations, while the plum produced intermediate force levels with gradual slopes, reflecting its natural compliance and internal damping. The data plots for these two softer cases can be found in Appendix G.\\
\noindent\hspace{1em}The average firmness values computed from multiple trials are plotted in Figure~\ref{fig:21.Firmness}. These results demonstrate the system’s potential to distinguish objects of varying stiffness, validating the feasibility of real-time, in-hand firmness assessment for automated sorting and quality grading applications in the fruit industry.
\subsection{Firmness Estimation on Apricots} 
\subsubsection{Progressive Firmness Monitoring Test}
\begin{figure}[h!]
    \centering
    \includegraphics[width=0.8\linewidth]{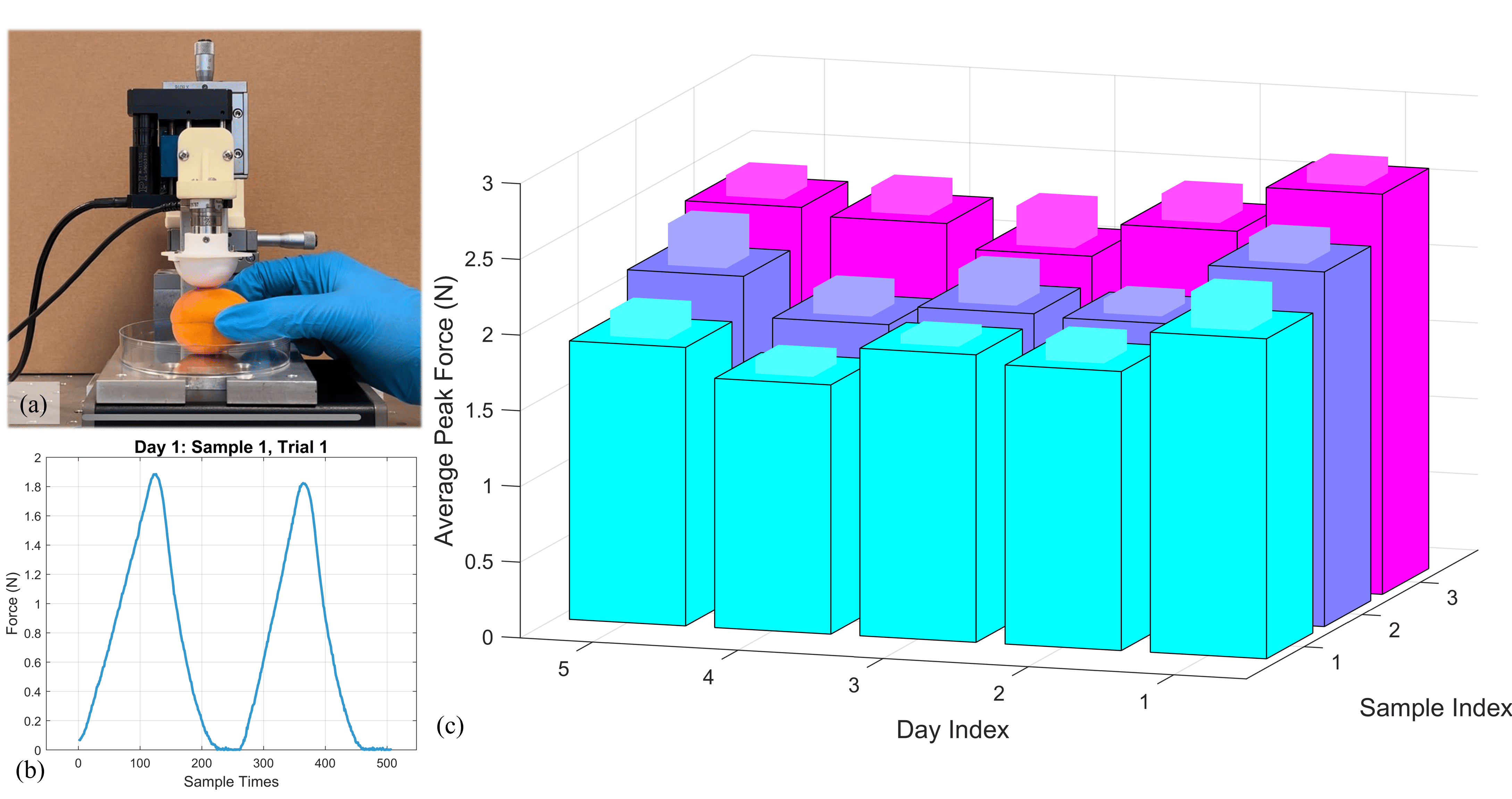}
    \caption{Reference indentation results in the progressive firmness test: (a) Close-up of the indentation setup; (b) Example Day 1–Sample 1–Trial 1 force data; (c) Average peak indentation forces across samples and days.}
    \label{fig:22.DatingRef}
\end{figure}

\begin{figure}[h!]
    \centering
    \includegraphics[width=0.8\linewidth]{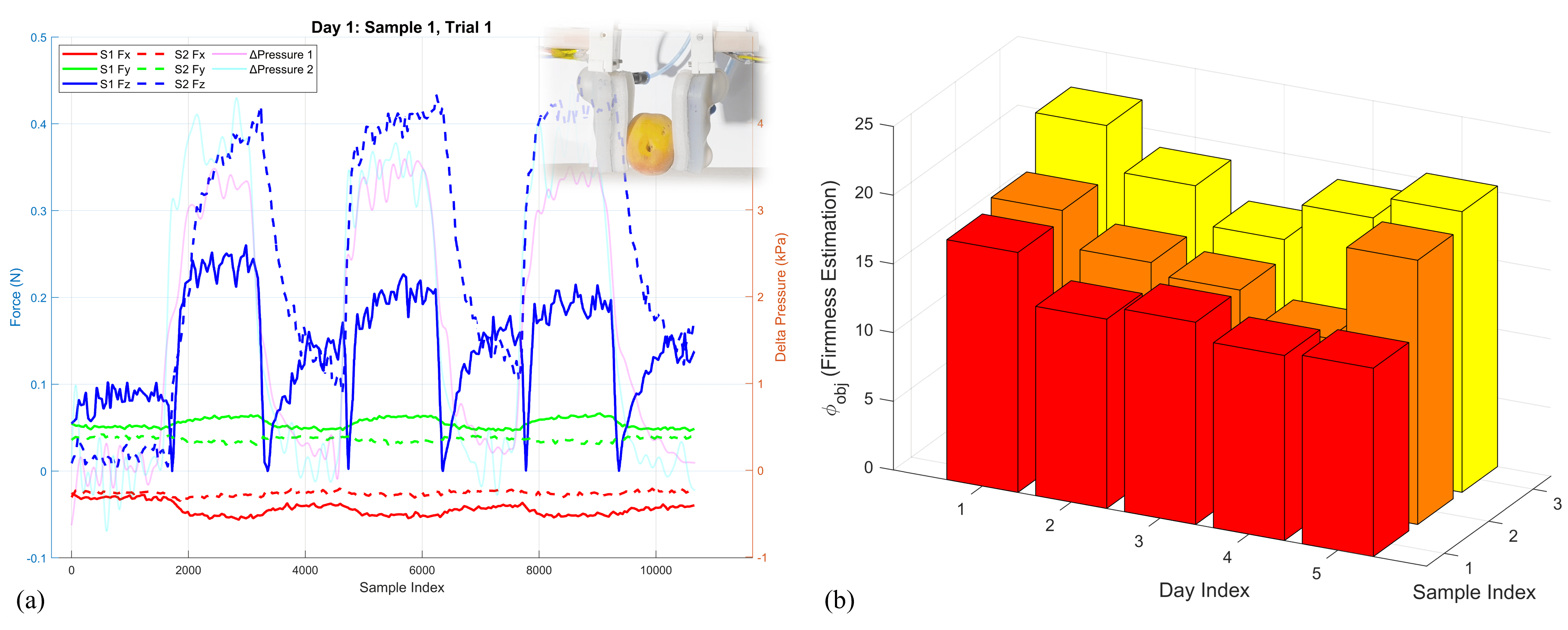}
    \caption{Progressive firmness evaluation using SoftMag gripper: (a) Representative data for Day 1–Sample 1–Trial 1 probing; (b) Average firmness estimation across samples and days.}
    \label{fig:23.Firmness}
\end{figure}

\noindent\hspace{1em} To further assess the system’s capability for tracking fruit ripeness, a five-day progressive firmness test was conducted on three apricots. As a reference, indentation tests were performed using the same ABS probe setup from Section 3.4. Three apricots, purchased from a local supermarket, were selected as the testing objects. Each fruit was tested daily with five indentation trials, following a consistent orientation protocol to ensure repeatability. Each trial consisted of two cycles of indentation performed at a loading speed of 1.5 mm/s, with a fixed displacement of 2 mm and a sampling rate of 100 Hz. On the first day, the samples were labeled based on the data collected (the averaged peak indentation force), from soft to hard, as apricots 1, 2, and 3. The collected force data, forming a 5 × 3 × 5 dataset, were averaged by cycles and trials to get a 5 × 3 matrix. Figure~\ref{fig:22.DatingRef} (a) and (b) depict the test scene and a Day 1–Sample 1–Trial 1 reference data as an example, while Figure~\ref{fig:22.DatingRef} (c) presents the 3D bar plot of average firmness across days and samples with standard deviations. It is observed that from the sample index dimension, a consistent firmness-increasing trend is evident, with samples ranging from soft to hard (apricot 1 to apricot 3) across all five days. From the time dimension, firmness generally decreases over the first 3–4 days, followed by an increase during the last 1–2 days. These trends align with the known physiology of post-harvest softening followed by dehydration-induced tissue stiffening.\\
\noindent\hspace*{1em}In parallel, the same apricots were evaluated daily using the SoftMag gripper. For each fruit, two probing trials were performed per day, each including three pressure cycles. A representative data plot (Figure~\ref{fig:23.Firmness} (a)) shows the probing response for apricot 1 on Day 1. The collected force responses were used to estimate firmness values via Equation (6.1). The results also formed a 5 × 3 matrix as shown in Figure~\ref{fig:23.Firmness} (b). Notably, to prevent actuator failure during the progressive testing, only two trials were conducted for each condition. As a result, the standard deviation is not reported in the firmness estimation result.\\

\noindent\hspace{1em} To quantify the agreement between the proposed probing-based method and the standard indentation setup, a Pearson correlation analysis was performed across both the sample and time dimensions\cite{blanke2013non}. The analysis involved calculating the Pearson correlation coefficient ($r$), coefficient of determination ($r^2$), $p$-value, and the 95\% confidence interval (CI) derived from the standard error. Full methodological details are reported in Appendix~G. The results for all four statistical indices are summarized in Table~1 and Table~2. In the sample dimension, Pearson correlation coefficients range from $0.818$ to $0.920$, indicating a strong positive linear relationship between the reference and SoftMag-based firmness evaluations for all three samples. The coefficients of determination ($r^2$) show that $65.7\%$ to $84.6\%$ of the variance in the indentation-based reference data can be explained by the gripper's predictions. Statistically significant correlations ($p < 0.05$) were observed for Sample~1 and Sample~3, with $p$-values of $0.031$ and $0.027$, respectively. While Sample~2’s $p$-value of $0.096$ exceeds the conventional threshold, the result still reflects a moderately strong correlation. The confidence interval half-widths ranged from $0.445$ to $0.662$, with narrower intervals corresponding to samples with stronger correlations, indicating reasonably high reliability in the estimates. In the time dimension, the $r$ values range from $0.889$ to $0.978$, confirming a consistently strong linear association between the two evaluation methods across all days. The $r^2$ values, from $78.9\%$ to $95.6\%$, indicate that the proposed method captures most of the temporal variability observed in the reference tests. While the $p$-values ($0.135$–$0.304$) exceed the conventional significance threshold, this is attributable to the small number of temporal samples and does not contradict the consistently high $r$ and $r^2$ values, which point to a strong underlying relationship. The wider confidence intervals ($0.413$–$0.900$) reflect lower statistical precision in the time domain compared to the sample domain, yet the trends remain consistent and interpretable. Collectively, these results support the ability of the SoftMag gripper to track firmness variations accurately and reliably across fruit samples and throughout the ripening timeline.

\begin{table}[htbp] 
    \centering
    \caption{Correlation analysis for the progressive test in the sample dimension.}
    \label{tab:sample_corr}
    \begin{tabular}{l
            S[table-format=1.3]
            S[table-format=1.3]
            S[table-format=1.3]
            S[table-format=1.3]}
    \toprule
     & {$r$} & {$r^2$} & {$p$-value} & {CI half-width} \\
    \midrule
    Sample 1 & 0.912 & 0.832 & 0.031 & 0.463 \\
    Sample 2 & 0.818 & 0.657 & 0.096 & 0.662 \\
    Sample 3 & 0.920 & 0.846 & 0.027 & 0.445 \\
    \bottomrule
    \end{tabular}

    \vspace{2ex} 
    \centering
    \caption{Correlation analysis for the progressive test in the time dimension.}
    \label{tab:time_corr}
    \begin{tabular}{l
            S[table-format=1.3]
            S[table-format=1.3]
            S[table-format=1.3]
            S[table-format=1.3]}
    \toprule
     & {$r$} & {$r^2$} & {$p$-value} & {CI half-width} \\
    \midrule
    Day 1 & 0.961 & 0.923 & 0.179 & 0.544 \\
    Day 2 & 0.978 & 0.956 & 0.135 & 0.413 \\
    Day 3 & 0.968 & 0.938 & 0.161 & 0.490 \\
    Day 4 & 0.889 & 0.789 & 0.304 & 0.900 \\
    Day 5 & 0.934 & 0.872 & 0.233 & 0.703 \\
    \bottomrule
    \end{tabular}
\end{table}

\subsubsection{Individual Firmness Estimation Test}
\begin{figure}[h!]
    \centering
    \includegraphics[width=0.9\linewidth]{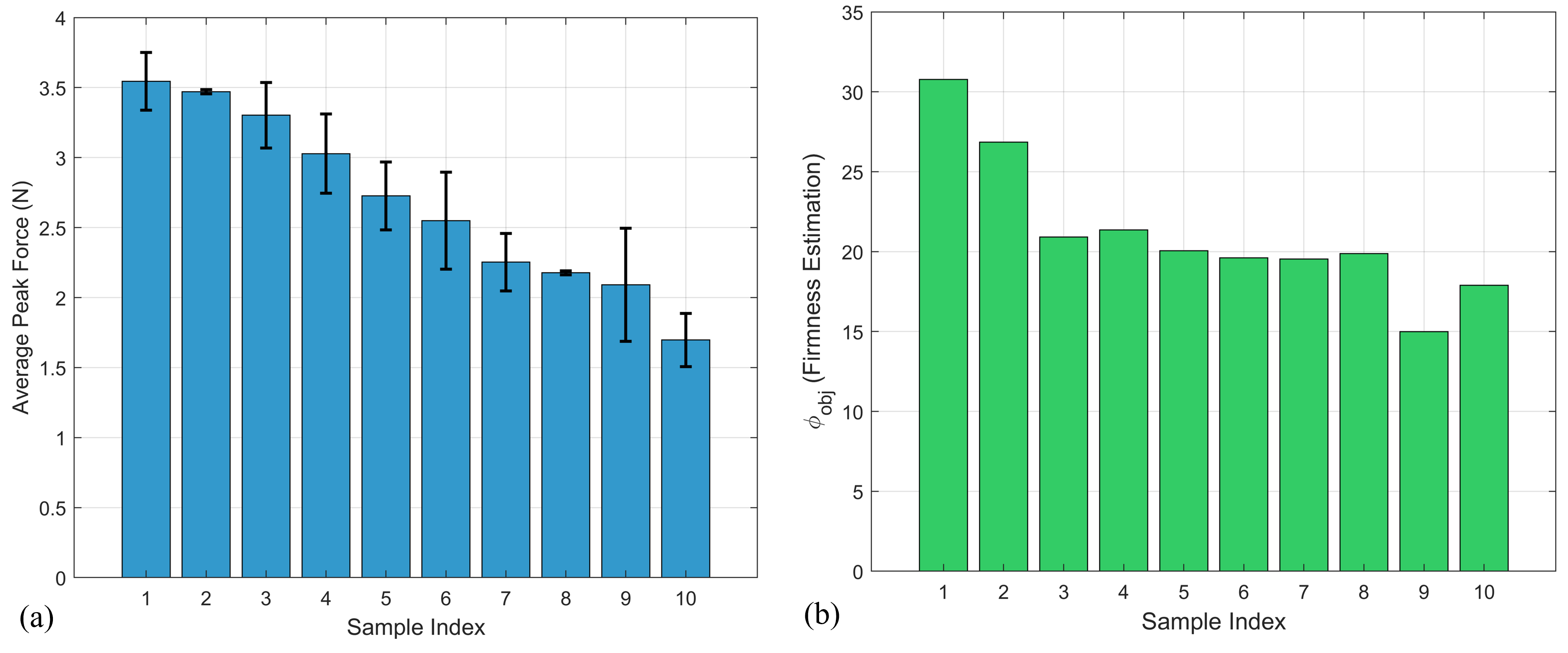}
    \caption{Individual firmness evaluation results across ten apricot samples: (a) Average peak forces (reference); (b) Firmness estimations by the SoftMag firmness evaluation framework.}
    \label{fig:24.Apricot}
\end{figure}
\noindent\hspace{1em}To complement the progressive test, a time-independent individual firmness evaluation was conducted to further validate the system’s performance. Ten apricots were first labeled arbitrarily and tested using the SoftMag gripper since the probing had negligible impact on fruit integrity. Subsequently, the same samples were evaluated using the reference indentation setup from Section 3.4. Peak forces were averaged across cycles and trials, with standard deviations shown in Figure~\ref{fig:24.Apricot} (a). Based on these results, the samples were relabeled from softest to firmest (1–10), and the corresponding firmness estimates from the SoftMag system were plotted in Figure~\ref{fig:24.Apricot} (b). Again, standard deviations for the SoftMag data are omitted due to the limited number of trials performed to avoid actuator failure. The same correlation analysis was applied to assess the consistency between the proposed evaluation system and the reference results. The analysis revealed a coefficient of $r=0.829$, confirming a strong positive linear relationship between the two datasets. The $r^2$ value of 0.687 indicates that $68.7\%$ of the variance in the reference data is captured by the SoftMag system. A p-value of 0.003 confirms statistical significance, while the CI half-width of 0.388 reflects reasonable estimation precision. Together with the progressive test, these results further demonstrate the reliability and applicability of the proposed probing-based method for non-destructive, in-hand firmness evaluation across fruit samples.

\section{Discussion and Future Work}
\noindent\hspace{1em}This study presents a unified framework for designing, fabricating, and evaluating a soft robotic gripper with integrated magnetic tactile sensing. By combining shared-material integration, signal decoupling, and learning-based inference, the system enables real-time adaptive grasping and soft probing for firmness evaluation. A key contribution is the identification and mitigation of the mechanical parasitic effect—a long-overlooked issue in sensorized actuators that can distort tactile signals during actuation. Despite promising results, several challenges remain. The current spatial resolution, limited by a single Hall-effect sensor and sparse magnet layout, is insufficient for fine-grained contact mapping. Fabrication inconsistencies, such as variations in foam porosity and magnet alignment, introduce variability across actuators. While calibration helps compensate for these differences, future work should be focused on improving manufacturing consistency or exploring self-calibrating mechanisms. Environmental sensitivity, especially to nearby ferromagnetic materials, also poses limitations in unstructured settings, though our interference tests help define safe operating margins. It is worth noting that the mechanical parasitic effect studied in this work, while undesirable in applications requiring tactile sensing (e.g., contact force estimation), may be beneficial in scenarios where proprioception or shape sensing is of interest. Another important consideration is application suitability. While the system successfully tracked firmness changes in apricots, climacteric fruits like kiwis or avocados, with clearer stiffness trends, may provide better test cases. Moreover, the added industrial value of the system depends on the market value of the target fruit. High-value crops such as grapes, particularly in wine production, may offer more compelling use cases. Looking ahead, we are progressing toward an upgraded tri-finger version of the SoftMag gripper to allow stable grasps of objects larger (and irregularly shaped) than what has been tested so far, as well as their effective probing. Future work will focus on robotic arm integration for autonomous sorting, expanding sensor coverage across the actuator surface, and incorporating hybrid modalities for richer feedback. Achieving high-resolution multitouch sensing will be critical for advancing the system’s capabilities in complex tactile interactions. Vision-based modules will also be integrated to support both grasp planning and ripeness evaluation. Meanwhile, a preliminary demonstration of adaptive control to avoid slippage using real-time sensing data is shown in Video 7, illustrating the potential for closed-loop strategies to improve grasp robustness. Broader applications, such as soft object inspection in food, may also benefit from the system’s compliant and material-aware sensing capabilities.

\section*{Conclusion}
\noindent\hspace{1em}This work presents the development of the SoftMag system, a soft robotic actuator (thus gripper) that integrates magnetic-based tactile sensing with pneumatic actuation for adaptive grasping and in-hand firmness evaluation. Starting from the design and fabrication of a sensorized actuator, a complete framework encompassing multiphysics simulation, systematic characterization, and system-level integration was established. The mechanical parasitic effect that prevails in soft tactile-driven actuators was explicitly identified and resolved using a neural-network-based decoupling method. A multi-task learning model was employed for real-time prediction of force and contact position, while a probing-based strategy enabled continuous, quantitative firmness estimation. The system was validated through grasping experiments and fruit evaluation tests, demonstrating its potential for non-destructive, in-process quality assessment. Magnetic interference studies further guided the safe operational range of the sensor. Altogether, this work contributes a unified approach to the design and characterization of soft tactile-based actuator/grippers. The proposed SoftMag gripper system provides a foundation for future deployment in applications such as fruit sorting, packaging, and soft object inspection, where gentle yet informed handling is critical.

\section*{Declaration of conflicting interests}
\noindent\hspace{1em}The author(s) declared no potential conflicts of interest with respect to the research, authorship, and/or publication of this article.

\section*{Funding}
\noindent\hspace{1em}The author(s) disclosed receipt of the following financial support for the research, authorship, and/or publication of this article: The research was partially funded by the European Union - NextGenerationEU and by the Ministry of University and Research (MUR), National Recovery and Resilience Plan (NRRP), Mission 4, Component 2, Investment 1.5, project “RAISE - Robotics and AI for Socio-economic Empowerment” (ECS00000035).


\end{document}